\documentclass[journal]{IEEEtran}

\usepackage{times}
\usepackage{xcolor}
\usepackage{soul}
\usepackage[utf8]{inputenc}
\usepackage{caption}
\usepackage[pdftex]{graphicx}
\usepackage{amsmath,amsfonts,amsthm,amssymb}
\usepackage[CJKbookmarks=true,
            bookmarksnumbered=true,
            bookmarksopen=true,
            colorlinks=true,
            citecolor=green,
            linkcolor=red,
            anchorcolor=black,
            urlcolor=blue
            ]{hyperref}
\usepackage[flushleft]{threeparttable}
\usepackage{array}
\usepackage{subfigure}
\usepackage{booktabs, caption}
\usepackage{float}
\begin{document}

\title{Asymmetric GAN for Unpaired \\ Image-to-image Translation}

\author{Yu~Li,
Sheng~Tang,~\IEEEmembership{Member,~IEEE,}
Rui~Zhang,
Yongdong Zhang, ~\IEEEmembership{Senior Member,~IEEE,}
Jintao Li, ~\IEEEmembership{Member,~IEEE,}
and~Shuicheng Yan,~\IEEEmembership{Fellow,~IEEE}

\IEEEcompsocitemizethanks{\IEEEcompsocthanksitem Copyright (c) 2018 IEEE.

Manuscript received July 17th, 2018; revised Jan 7th, 2019; resubmitted Feb 26th, 2019; minor revised Apr 30th, 2019; accepted June 4th, 2019.

This work was supported in part by the National Key Research and Development Program of China under Grant 2017YFC0820605, in part by the National Natural Science Foundation of China under Grant 61572472, Grant 61871004, and Grant 61525206, and in part by 242 Project under Grant 2019A010. \emph{(Corresponding author: Sheng Tang.)}

Y. Li and R. Zhang are with the Key Laboratory of Intelligent Information Processing, Institute of Computing Technology, Chinese Academy of Sciences, Beijing 100190, China, and also with the School of Computer Science and Technology, University of Chinese Academy of Sciences, Beijing 100049, China (e-mail: liyu@ict.ac.cn; zhangrui@ict.ac.cn).

S. Tang, Y. Zhang, and J. Li are with the Key Laboratory of Intelligent Information Processing, Institute of Computing Technology, Chinese Academy of Sciences, Beijing 100190, China (e-mail: ts@ict.ac.cn; zhyd@ict.ac.cn; jtli@ict.ac.cn).

S. Yan is with the AI Institute, Qihoo 360, Beijing 100025, China, and also with the Department of Electronics and Computer Engineering, National University of Singapore, Singapore (e-mail: yanshuicheng@360.cn).
}
\thanks{}}
\markboth{}
{}
\maketitle

\begin{abstract}
Unpaired image-to-image translation problem aims to model the mapping from one domain to another with unpaired training data.
Current works like the well-acknowledged Cycle GAN provide a general  solution for any two domains through modeling injective mappings with a symmetric structure.
While in situations where two domains are asymmetric in complexity, \emph{i.e.} the amount of information between two domains is different, these approaches pose problems of poor generation quality, mapping ambiguity, and model sensitivity.
To address these issues, we propose Asymmetric GAN (AsymGAN) to adapt the asymmetric domains by introducing an \emph{auxiliary variable (aux)} to learn the extra information for transferring from the information-poor domain to the information-rich domain, which improves the performance of state-of-the-art approaches in the following ways.
First, \emph{aux} better balances the information between two domains which benefits the quality of generation.
Second, the imbalance of information commonly leads to mapping ambiguity, where we are able to model one-to-many mappings by tuning \emph{aux}, and furthermore, our \emph{aux} is controllable.
Third, the training of Cycle GAN can easily make the generator pair sensitive to small disturbances and variations while our model decouples the ill-conditioned relevance of generators by injecting \emph{aux} during training.
We verify the effectiveness of our proposed method both qualitatively and quantitatively on asymmetric situation, label-photo task, on Cityscapes and Helen datasets, and show many applications of asymmetric image translations. 
In conclusion, our AsymGAN provides a better solution for unpaired image-to-image translation in asymmetric domains.
\end{abstract}


\IEEEpeerreviewmaketitle

\begin{figure*}
  \begin{center}
    \includegraphics[width=\linewidth]{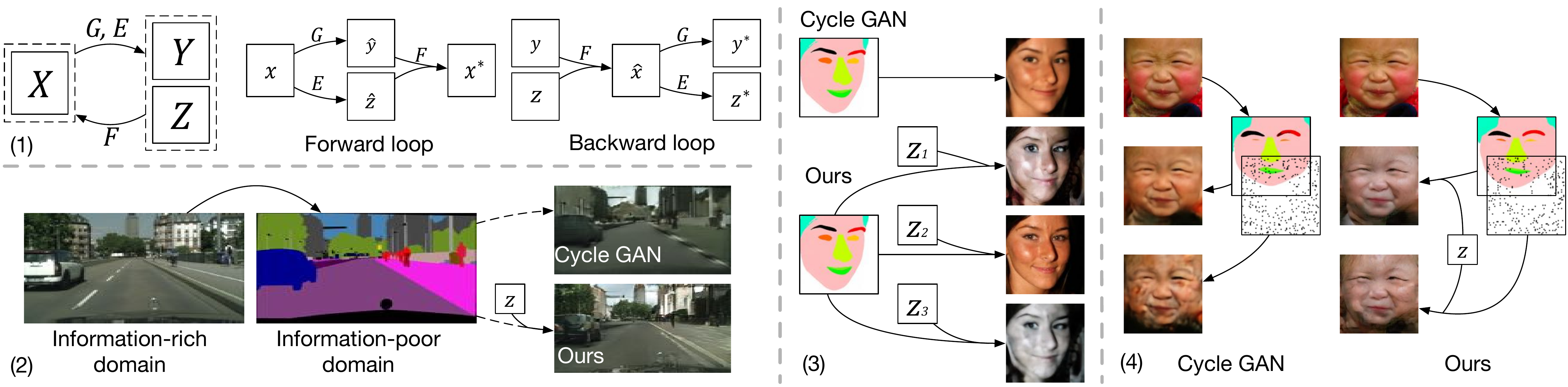}
    \caption{(1) An overview of AsymGAN. (2) Information lost when generating a label image from a photo, so it is hard to recover a photo from the label image. After adding \emph{aux}, a better result is obtained. (3) The mapping of label$\rightarrow$photo should be one-to-many. Diverse outputs can be obtained with different \emph{aux}es. (4) The model sensitivity problem. Cycle GAN recovers $x$ with $F(G(x))$ perfectly, but is seriously influenced by small disturbances, and $F(G(x)+\epsilon)$ is bad. Our model performs more robust.}
    \vspace*{-20pt}
    \label{fig_motivation}
  \end{center}
  \end{figure*}

\section{Introduction}
\label{sec_introduction}
\IEEEPARstart{I}{mage-to-image} translation is a series of vision and graphics problems which is required to map an image from one domain to another.
Pix2pix \cite{isola2016image} develops a common framework as a general solution for paired image-to-image problems.
Zhu \emph{et al.} \cite{zhu2017toward} further proposes Bicycle GAN to model a distribution of possible outputs in situations where the mappings between two domains are ambiguous.
Some other approaches \cite{lu2017sketch,taigman2016unsupervised,kim2017learning} design frameworks for specific tasks.

On the other hand, in conditions where aligned image pairs are not available, Cycle GAN \cite{zhu2017unpaired} presents an ingenious symmetric structure to learn mappings both from $X$ to $Y$ and $Y$ to $X$ with a cycle consistency loss to enforce $F(G(X))\approx X$ and $G(F(Y))\approx X$ where $G$ and $F$ are mapping functions.
This symmetric structure in Cycle GAN manages to model a pair of reciprocal generators between two domains to generate faithful images in many tasks such as horse$\leftrightarrow$zebra and season transfer, where the complexity of the two domains are at the same level.
However, when translating between domains where complexity disparity exists, namely the information of the two domains is asymmetric and imbalanced such as labels$\leftrightarrow$photos, the following  problems arose:
\begin{enumerate}
  \item \textbf{Poor quality:} The imbalanced information of the two domains leads to poor quality of generated images since it is naturally improper to require the generator to produce high complexity images with low complexity inputs.
  \item \textbf{Mapping ambiguity:} When translating an image from the information-poor domain to the information-rich domain, it is common that there should be more than one proper alternatives because many degrees are free to change.
  \item \textbf{Model sensitivity:} No matter how different the two domains are, to meet the cycle consistency loss during training, Cycle GAN model is easily forced to some ill-conditioned point where the model encodes the information of different domains in some invisible ways, which distracts the model from high-quality translation and makes the model sensitive to small disturbances and variations of the inputs.
  We call this phenomenon model sensitivity problem.
\end{enumerate}

Our proposed Asymmetric GAN (AsymGAN) aims to solve these problems by introducing an \emph{auxiliary variable (aux)}, which is enforced to follow a specific distribution such as Gaussian, to encode the lost information when transferring images from information-rich domain to information-poor domain.
As a result, in reverse, while transferring images from information-poor domain to information-rich domain, the \emph{aux} is able to provide more information, which benefits training of both generators.
Fig.\ref{fig_motivation}(1) is a general overview of our method which has the following advantages.

First, with \emph{aux}, the complexity between two domains is better balanced, which enables the generation quality to be improved.
As illustrated in Fig.\ref{fig_motivation}(2), in situations like semantic labels$\leftrightarrow$photos, the information in real photos ($X$) is much more than that of semantic label maps ($Y$).
To construct a cycle, Cycle GAN is required to map a label image with inadequate information to a real photo full of details, which does no good to the improvement of generation quality due to the information imbalance.
On the contrary, we want to preserve information produced during translating a real photo $x$ to its corresponding $\hat{y}$ by encoding this information to a given distribution $p(z)$.
As a result, when we generate $\hat{x} = F(y, z)$ with $z \sim p(z)$, this sampled noise term is actually endowed with extra informations learned from $X\rightarrow \hat{Y}$.
We complement the information for both $G$ and $F$ by \emph{aux}, which benefits training of both $G$ and $F$ resulting in higher quality of the translated images since $G$ and $F$ are highly correlated and are improved simultaneously.

Second, we model a distribution of the output alternatives conditioned on the input rather than model an injective function.
Intuitively, mappings like label$\rightarrow$photo, sketch$\rightarrow$photo should be one-to-many because the asymmetric domains provide many free degrees during transferring.
The symmetric structure of Cycle GAN has constrained its mappings to be one-to-one.
Consequently, $G$ and $F$ are both injective functions focusing on generating a single result conditioned on the input.
However, in AsymGAN, as shown in Fig.\ref{fig_motivation}(3), it is natural to obtain various output images from one input by utilizing different \emph{aux}es.
Moreover, we can not only sample different \emph{aux}es from the priori distribution $p(z)$, we can also control the generation diversity by utilizing encoded \emph{aux}es of other $x$es ($x \in X$).

Third, our method can alleviate the sensitivity problem of Cycle GAN.
The sensitivity problem is that Cycle GAN is easily converged to some state where the models $G$ and $F$ are sensitive to small disturbances or variations and hence ill-conditioned.
Take the case of the forward loop $x\rightarrow G(x) \rightarrow F(G(x))$.
One of its objective is to minimize $\mathbb{E}_{x\sim p_{data}(x)}||F(G(x))-x||_1$ where $x\in X$.
It is observed that $F(G(x))$ can perfectly recover $x$ in almost all the cases of $G(x)$ even $x$ is not in training set.
However, when applying $F$ to some real data $y$, the result is far from satisfied.
Even some small disturbances $\epsilon$ or variations on $G(x)$ could lead to a much weaker result.
Fig.\ref{fig_motivation}(4) shows the influence of adding disturbances $F(G(x)+\epsilon)$.
The reason of this sensitivity problem may be that $G$ and $F$ encode the information of $X$ and $Y$ in some extreme and trivial ways to cater to the cycle consistency losses precisely, which leads to ill-conditioned model.
In our case, since we continuously inject \emph{aux} $z\sim p(z)$, which can also be considered as a noise term, in $F(G(x), z)$ to decouple the trivial relevance between $G$ and $F$, we are able to alleviate this sensitivity problem (as shown in Fig.\ref{fig_motivation}(4)).
In another view, we provide another container \emph{aux} for the generators so that they have another choice to encode the information rather than encode it in some trivial ways.
Since our structure helps the model to concentrate on \emph{generating} instead of \emph{encoding and decoding the lost information in some secret ways}, the generation quality of both directions could be further improved.

The main contributions of this paper are as follows:
\begin{itemize}
  \item We conclude the problems of Cycle GAN when applying on asymmetric domains, among which, we observe the sensitivity problem and report the influence of small translation, scale transformation, and disturbances.
  \item We propose an Asymmetric GAN framework to model the unpaired image-to-image translation between asymmetric domains.
  \item Experiments verify that between asymmetric domains, our AsymGAN is able to generate images of better quality, produce diverse outputs, and alleviate the sensitivity convergence problem.
\end{itemize}

\section{Related Work}
With the rapid development of machine learning, neural networks are largely explored to build discriminative models and impressive progresses have been made in many fundamental computer vision problems such as image classification \cite{krizhevsky2012imagenet, lin2013network, simonyan2014very, szegedy2015going, he2016deep}, object detection \cite{girshick2016region, girshick2015fast, ren2015faster, lin2017focal, li2018implicit}, segmentation \cite{Yu2016Multi, chen2016deeplab,liang2016semantic, ZhangR:SP:IJCAI17, ZhangR:SP:ICCV17, he2017mask}, and image caption \cite{karpathy2015deep, li2017image, li2018gla}.
It is not until the proposal of generative adversarial networks (GANs) \cite{goodfellow2014generative} that learning generative models with deep neural networks achieves reasonable results and draw lots of research attention.
Image-to-image translation can be considered as a general application of GAN.
Thus, we introduce generative adversarial networks first, and then image-to-image translation method, and finally discuss the approaches that use latent vectors and the challenges of using latent vectors in generative models.

\subsection{Generative Adversarial Networks}
As a class of the most successful generative models for photorealistic image generation, GANs intend to learn a pair of generator and discriminator from the min-max game.
The discriminator learns to distinguish the real and fake images, while the generator learns to map the random noises sampled from a known distribution to plausible images and fool the discriminator.
Due to the unstable training problem of the original GAN \cite{goodfellow2014generative}, approaches like WGAN \cite{arjovsky2017wasserstein,gulrajani2017improved} and loss-sensitive GAN \cite{qi2017loss} are proposed to stabilize training.

Meanwhile, conditional GANs (cGANs) \cite{mirza2014conditional} have also been actively studied and successfully applied to many tasks.
Some of these generate images conditioned on discrete labels \cite{mirza2014conditional} or text \cite{zhang2016stackgan,zhang2017stackgan++}.
Image-to-image translations usually generate target images conditioned on input images.

In this paper, we take the advantages of GANs to synthesize plausible images and utilize the conditional least squares GAN \cite{mao2017least} as the basic structure of our framework.

\subsection{Image-to-image Translation}

Beyond the framework of cGAN, a large variety of challenging image-to-image tasks have been tackled, among which image inpainting \cite{pathak2016context, mirza2014conditional}, super-resolution \cite{ledig2016photo}, age progression and regression \cite{zhang2017age}, face attribute manipulation \cite{shen2017learning,kim2017learning,zhou2017genegan}, scene synthetic \cite{wang2016generative, zhang2018style}, makeup applying and removing \cite{chang2018pairedcyclegan}, style transfer \cite{zhu2017unpaired} , \emph{etc.}, are of significant representativeness.

Most of the approaches mentioned above are specifically designed for particular applications and are not general.
Pix2pix \cite{isola2016image} first manages to develop a common framework for all problems requiring image pairs for training.
It combines an adversarial loss along with a $l_1$ loss to learn these tasks in a supervised manner using cGANs, thus requires paired samples.
Bicycle GAN \cite{zhu2017toward} improves Pix2pix by modeling a distribution of potential results, as many of these problems may be multi-modal or ambiguous in nature.
Although they are able to produce various outputs by sampling different latent vectors, the output diversity of the generated images is beyond control.

To release the burden of obtaining data pairs, unpaired image-to-image translation frameworks \cite{liu2017unsupervised,kim2017learning,yi2017dualgan,zhu2017unpaired} have been proposed.
UNIT \cite{liu2017unsupervised} is a combination of variational autoencoders (VAEs) \cite{kingma2013auto} and CoGAN \cite{liu2016coupled}. DiscoGAN \cite{kim2017learning}, DualGAN \cite{yi2017dualgan} and Cycle GAN \cite{zhu2017unpaired} share exactly the same structure and employ a cycle consistency loss to preserve key attributes between the input and output images.

Recent work AugCGAN \cite{Almahairi2018AugmentedCL} proposes a similar framework compared with ours at first sight. 
However, AugCGAN is designed to model many-to-many mappings between two domains substantially, by introducing $z_a$ and $z_b$ to explicitly encode the difference of domain $a$ and $b$ ($\tilde{b} = G_{AB}(a, z_b) , \tilde{z_b} = E_B(b, \tilde{a})$, vice versa).
Thus, during inference, AugCGAN is able to generate diverse outputs $\tilde{a}$ and $\tilde{b}$ by sampling different $z_a\sim p(z_a)$ and $z_b\sim p(z_b)$.
Our AsymGAN is motivated to adding $z$ to provide another path for information flow between asymmetric domains to obtain better generation quality ($\hat{x} = F(y, z), \hat{z} = E(x')$).
Additionally, the added auxiliary variable enables us to model multi-modality mappings.
Furthermore, to produce diverse outputs, in addition to utilizing sampled $z$ from $p(z)$, utilizing encoded $z$ enables us to control the output diversity since our $z$ can be encoded with $x'$ only.

On the other hand, starGAN \cite{choi2017stargan} aims to perform image-to-image translations for multiple domains by using one model.

In this paper, we mainly discuss the two-domain translation problem.
Though Cycle GAN \cite{zhu2017unpaired} achieves state-of-the-art performances and is considered to be a solid baseline, it suffers from the three problems as aforementioned in Section \ref{sec_introduction}.
Our AsymGAN aims at performing better in solving these problems.

\subsection{Generative Models with Latent Vectors}
\label{subsec_aux}
VAE \cite{kingma2013auto} attempts to encode images to latent vectors that follow a specific Gaussian distribution.
Thus, the network is able to decode any sampled latent vector to a real image.
Especially, VAE employs the reparameterization trick, which means that the latent vector is not directly generated by the encoder, but is sampled from a Gaussian distribution which is constructed by an encoded mean and deviation vector.
This distribution is constrained by KL divergence with normal distribution during training.

On the other hand, the original GAN \cite{goodfellow2014generative} and its subsequent works \cite{qi2017loss, arjovsky2017wasserstein, gulrajani2017improved} are focusing on modeling the mapping between a known distribution like Gaussian to an image domain distribution.
Thus a noise term, namely a latent vector is sampled from the known distribution as the input of generator.
Conditional GANs \cite{mirza2014conditional} follow the same idea and transform the sampled latent vectors to certain domain images conditioned on class labels, source images, or sentences.
Therefore, the latent vector and condition vector are both inputs of the conditional generator.
Image-to-image translation is a specific example of conditional GAN, that is conditioned on source domain images.
As a result, it is straightforward to take a latent vector and a condition image as the input of the generator.

Ideally, different latent vectors should correspond to diverse output images for both GAN and cGAN models.
However, observations show that mode collapse problem \cite{goodfellow2016nips} often occurs, which means that the generator learns to map several different latent vectors to the same output point.
Image-to-image conditional GANs have made a substantial improvement in the quality of the results, while the generator learns to largely ignore the random sampled latent vectors when conditioned on a relevant context \cite{zhu2017toward, isola2016image, pathak2016context, yang2017high, zhu2017unpaired}.
It has even been shown that ignoring the noise leads to more stable training \cite{zhu2017toward, isola2016image, pathak2016context}.
Consequently, general image-to-image translation methods such as Pix2pix \cite{isola2016image} and Cycle GAN \cite{zhu2017unpaired} take out the latent vector and utilize only the source domain images as inputs of the generator.
To attain multimodality, Bicycle GAN \cite{zhu2017toward} combines VAE \cite{kingma2013auto} with Pix2pix \cite{isola2016image} framework to force the latent vector produce diversity in paired situations.

Hence, how to make the latent vector produce diversity in paired image-to-image translation model is still a very challenging problem.
Furthermore, how to make the latent vector meaningful while maintaining the quality of output images in unpaired situations and being able to control the diversity is more ambitious.
As a result, our AsymGAN is carefully designed to make the latent vector work as is supposed,
which leads to a boost in the generation quality, output diversity, and model sensitivity.

\begin{figure*}
\begin{center}
	\includegraphics[width=\linewidth]{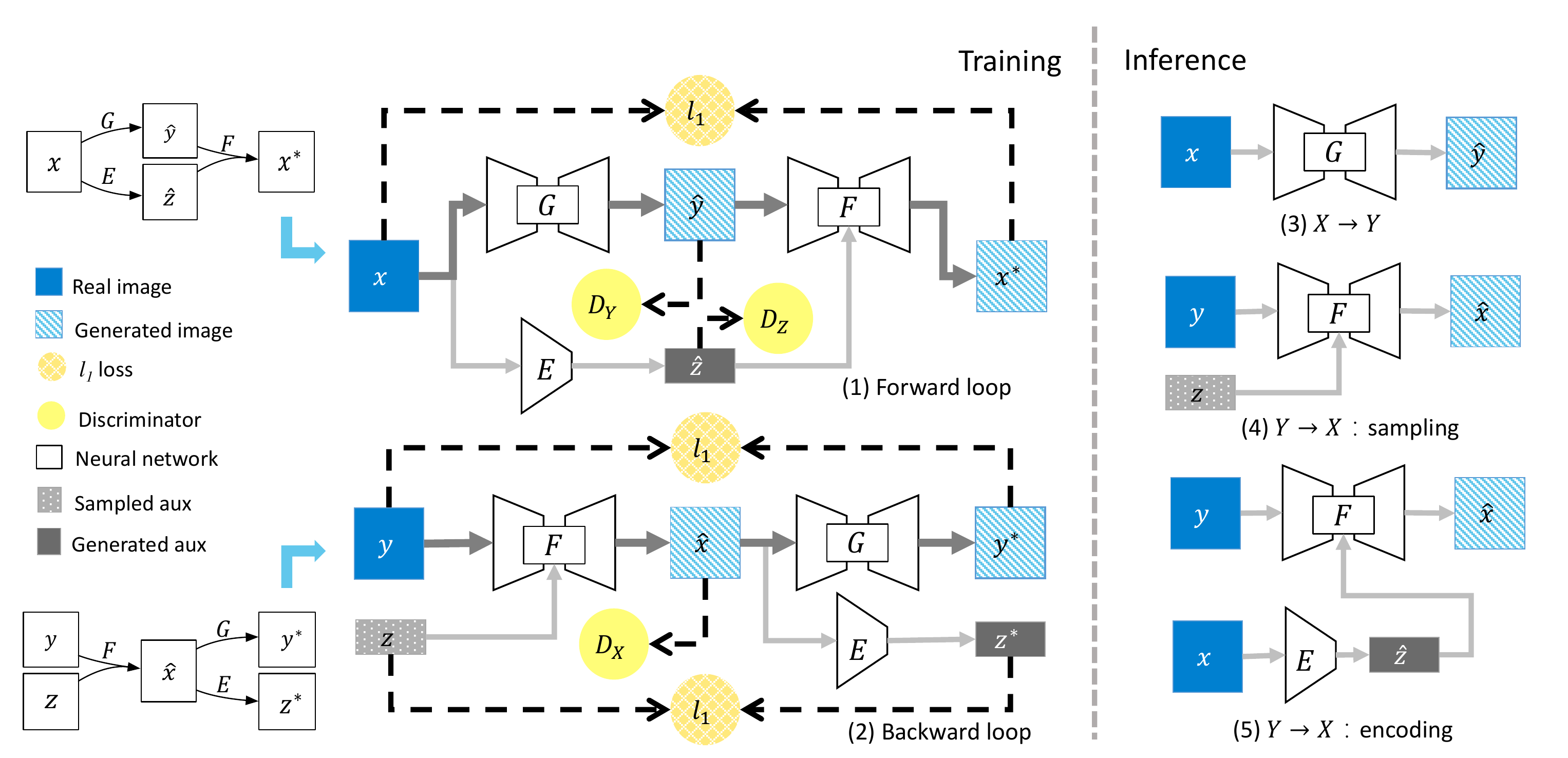}
	\caption{The main framework of Asymmetric GAN between information-rich $X$ and information-poor $Y$. (1) is the structure of the forward loop $x\rightarrow \hat{y} \rightarrow x^*$. (2) is the structure of the backward loop $y\rightarrow \hat{x} \rightarrow y^*$. (3)(4)(5) represent inference processes. (3) $\hat{y}=G(x)$. (4) Inference by sampling. We can generate $\hat{x}$ with a sampled $z$, $\hat{x}=F(y,z)$. (5) Inference by encoding. $\hat{x}$ can also be generated with an encoded $\hat{z}$, $\hat{x}=F(y,E(x))$.}
	\vspace*{-20pt}
	\label{fig_framework}
\end{center}
\end{figure*}

\section{Asymmetric GAN}

\label{sec_method}

We will first introduce our baseline model Cycle GAN, and then describe our proposed Asymmetric GAN.
Next, we summarize our full objective and detail our extension losses.

\subsection{Baseline: Cycle GAN}

Our goal is to learn mapping functions between two asymmetric domains $X$ and $Y$, where $X$ represents the domain that contains relatively rich information while $Y$ represents the domain that contains relatively poor information.
For example, in label$\leftrightarrow$photo task, $X$ represents domain ``photo" and $Y$ represents domain ``label".
Given unpaired training samples\footnote{The subscript $i,j$ are often omitted for simplicity.} $\{x_i\}_{i=1}^N$ and $\{y_j\}_{j=1}^M$ where $x_i\in X \subset \mathbb{R}^{H\times W\times 3}$ and $y_j\in Y \subset \mathbb{R}^{H\times W\times 3}$, we are going to model two mappings $G: X\rightarrow Y$ and $F: Y \rightarrow X$.
We denote the data distribution as $x\sim p_{data}(x)$ and $y\sim p_{data}(y)$.
In addition, we introduce adversarial discriminators $D_X$ and $ D_Y$ to distinguish real samples $x\sim p_{data}(x), y\sim p_{data}(y)$ and generated samples $\hat{x}=F(y,z), \hat{y}=G(x)$.

The objective of Cycle GAN contains two types of terms: adversarial losses and cycle consistency losses.
The adversarial losses match the generated image distribution to the target image distribution, which is formalized as follows:

\begin{align}
\begin{split}
\mathcal{L}_{A}(G, D_Y)&=\mathbb{E}_{y\sim p_{data}(y)}[\log D_Y(y)] \\
&+\mathbb{E}_{x\sim p_{data}(x)}[\log (1-D_Y(G(x)))],
\label{eq_LAG}
\end{split}
\end{align}
\begin{align}
\begin{split}
\mathcal{L}_{A}(F, D_X)&=\mathbb{E}_{x\sim p_{data}(x)}[\log D_X(x)] \\
&+\mathbb{E}_{y\sim p_{data}(y)}[\log (1-D_X(F(y)))].
\end{split}
\end{align}
The cycle consistency losses prevent the learned mappings $G$ and $F$ from contradicting each other \cite{zhu2017unpaired}:
\begin{align}
\begin{split}
\mathcal{L}_{CX}(G, F)&=\mathbb{E}_{x\sim p_{data}(x)}[||F(G(x))-x||_1],
\end{split}
\end{align}
\begin{align}
\begin{split}
\mathcal{L}_{CY}(G, F)&=\mathbb{E}_{y\sim p_{data}(y)}[||G(F(y))-y||_1].
\label{eq_lcy}
\end{split}
\end{align}
The full objective is:
\begin{align}
\begin{split}
\mathcal{L}(G,F,D_X,D_Y)&=\mathcal{L}_{A}(G, D_Y)
+\mathcal{L}_{A}(F, D_X)\\
&+\lambda \mathcal{L}_{CX}(G, F)
+ \lambda \mathcal{L}_{CY}(G, F).
\end{split}
\end{align}
Thus we aim to solve:
\begin{align}
\begin{split}
G^*, F^*=\arg \min_{G,F} \max_{D_X,D_Y} \mathcal{L}(G,F,D_X,D_Y).
\end{split}
\end{align}

As we can see, the structure of Cycle GAN is perfectly symmetric.
Therefore, problems arise in situations where two domains are asymmetric, which motivates us to develop AsymGAN.

\subsection{Asymmetric GAN}

As aforementioned, since domain $X$ and $Y$ are asymmetric, we intend to add an auxiliary variable $z\in Z \subset \mathbb{R}^{a\times b \times c}$ to balance the information volume and encapsulate the ambiguous aspects of the output mode that are not present in the input image.
To enable stochastic sampling, we hope $z$ to be drawn from some prior distribution $p(z)$; we use normal distribution $\mathcal{N}(0,I)$ in this work.
To capture the residual information in $X\rightarrow Y$, an encoder $E$ is required to map $E:X\rightarrow Z$.
Thus, our model includes three parts $G: X\rightarrow Y$, $E: X \rightarrow Z$, and $F:(Y,Z)\rightarrow X$.
Moreover, we introduce another adversarial discriminator $D_Z$ to distinguish real samples $z\sim \mathcal{N}(0,I)$ and generated samples $\hat{z}=E(x)$.

The training framework is illustrated in Fig.\ref{fig_framework}(1)(2).
During inference, for $X\rightarrow Y$, we generate $\hat{y}$ with $G(x)$ (Fig.\ref{fig_framework}(3)).
For $(Y,Z)\rightarrow X$, we can either sample an auxiliary variable $z$ from $p(z)$ (Fig.\ref{fig_framework}(4)) or utilize $\hat{z}$ which is encoded by $E(x)$ with some other samples $x\in X$ (Fig.\ref{fig_framework}(5)).
We are able to control the output diversity by utilizing this encoding configuration for inference.
Since the encoded \emph{aux} we use for generation contains some diversity features of the encoded source image, the generated image should also represent these features.
Thus, the encoding configuration enables us to control the output diversity by choosing specific encoded source images.
Although current works that focus on the mapping ambiguity of paired image translation problems such as Bicycle GAN \cite{zhu2017toward} are able to produce various outputs by sampling different latent vectors, the output diversity of the generated images is beyond control.
Our framework provides a novel approach that we can not only obtain diversity by sampling different $z$es, but also control the output diversity by encoding $z$es from other $x$es.

\subsubsection{Forward Loop}
In the forward loop ($x\rightarrow \hat{y} \rightarrow x^*$), the key idea of our approach is to preserve the information of $x$ that could be lost when generating $\hat{y}=G(x)$  in \emph{aux} $\hat{z}=E(x)$, so that when we need to go back to $x^*$, we can utilize $\hat{y}$ as well as $\hat{z}$ ($x^*=F(\hat{y}, \hat{z})$) to balance the information during translation, as shown in Fig.\ref{fig_framework}(1).
To access $z$ in the backward loop ($y\rightarrow \hat{x} \rightarrow y^*$) where the information are originally
insufficient, motivated by VAE \cite{kingma2013auto}, we enforce the encoded $\hat{z}=E(x)$ following a prior distribution $p(z)$.
That is, the distribution of residual information from $X$ to $Y$ will be encoded in the prior distribution $p(z)$.
Thus, we borrow the adversarial loss $\mathcal{L}_A(G,D_Y)$ in Eq.\ref{eq_LAG} to ensure the realness of $\hat{y}$, and make a constraint on $\hat{z}$ as :
\begin{align}
\begin{split}
\mathcal{L}_{A}(E, D_Z)&=\mathbb{E}_{z\sim p(z)}[\log D_Z(z)] \\
&+\mathbb{E}_{x\sim p_{data}(x)}[\log (1-D_Z(E(x)))].
\end{split}
\end{align}
Also, since $F$ takes both $\hat{y}$ and $\hat{z}$ as input, our cycle consistency loss will be modified to:
\begin{align}
\begin{split}
\mathcal{L}_{CX}(G, F)&=\mathbb{E}_{x\sim p_{data}(x)}[||F(G(x), E(x))-x||_1].
\end{split}
\end{align}

Different from VAE, our encoder directly models the latent vector $Z$ rather than models $\mu$ and $\sigma$; we choose the adversarial mechanism to close the gap of the distribution of $\hat{Z}$ and normal distribution rather than utilizing KL divergence.

\begin{figure*}[!htb]

\begin{center}
	\includegraphics[width=\linewidth]{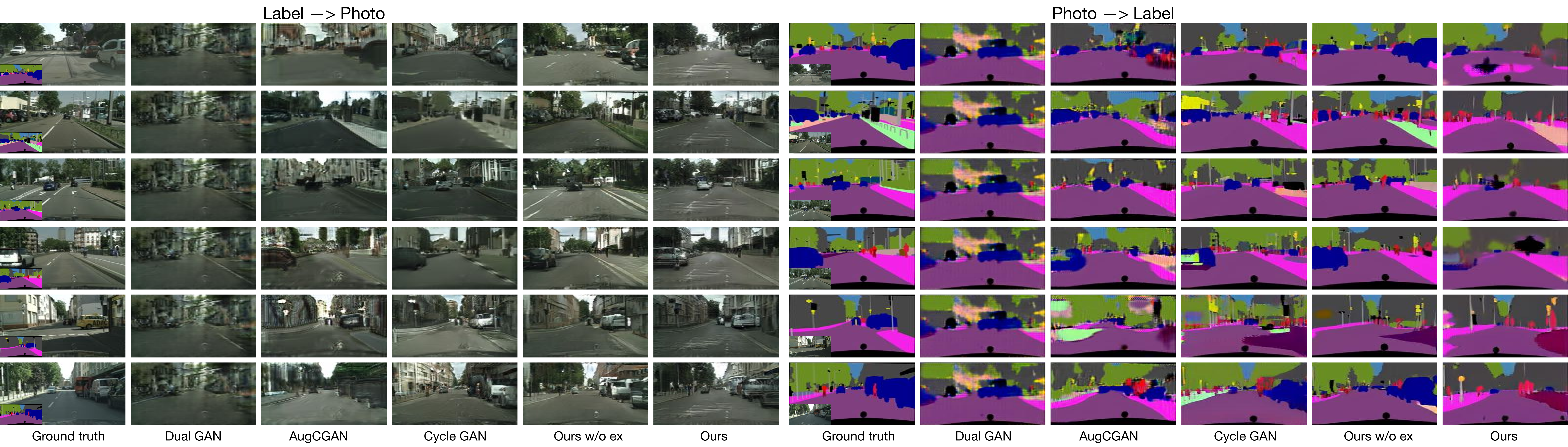}
	\caption{The comparison of generation quality on Cityscapes datasets between Cycle GAN and our method. The small picture on the left-bottom of ground truth image is the input image. Our approach identifies different categories clearly for both sides of mappings such as \emph{vegetation} and \emph{building} in the first row, and shows more details such as road lane line, lamps and windows of cars, \emph{etc.} for label$\rightarrow$photo.}
	\vspace*{-20pt}
	\label{fig_quality}
\end{center}
\end{figure*}

\subsubsection{Backward Loop}
In the backward loop ($y\rightarrow \hat{x} \rightarrow y^*$), as we already restrict the distribution of $\hat{Z}$, we can sample \emph{aux} from $p(z)$ to supplement information because different residual information has been mapped to different points on $p(z)$ as stated above, so $\hat{x}=F(y, z)$.
Furthermore, $\hat{x}$ will again produce an $z^*=E(\hat{x})$ when  generate $y^*=G(\hat{x})$.
To enhance the constraint on $E$ and avoid the problem of mode collapse on $z$, as claimed in \cite{isola2016image,zhu2017toward}, another cycle consistency loss for $z$ and $z^*$ is required to make sure that $Z$ can make a difference on the generation results.
As illustrated in Fig.\ref{fig_framework}(b), we first need an adversarial loss to pull the distribution of $\hat{x}$ to $p_{data}(x)$:
\begin{align}
\footnotesize
\begin{split}
\mathcal{L}_{A}(F, D_X)&=\mathbb{E}_{x\sim p_{data}(x)}[\log D_X(x)] \\
&+\mathbb{E}_{y\sim p_{data}(y),z\sim p(z)}[\log (1-D_X(F(y,z)))],
\end{split}
\end{align}
a cycle consistency loss on $y^*$ as in Eq.\ref{eq_lcy}:
\begin{align}
\footnotesize
\begin{split}
\mathcal{L}_{CY}(G, F)&=\mathbb{E}_{y\sim p_{data}(y),z\sim p(z)}[||G(F(y,z))-y||_1],
\end{split}
\end{align}
and another cycle consistency loss for $z^*$:
\begin{align}
\footnotesize
\begin{split}
\mathcal{L}_{CZ}(E, F)&=\mathbb{E}_{y\sim p_{data}(y),z\sim p(z)}[||E(F(y,z))-z||_1].
\end{split}
\end{align}
Thus, this cycle consistency loss on $z$ restricts the model from mode collapse and enforces the latent vector representing the missing dimensions of the information-poor domain.

\subsubsection{Full Objective}
\label{full_objective}
Above all, our full objective is:
\begin{align}
\footnotesize
\begin{split}
&\mathcal{L}(G,F,E,D_X,D_Y,D_Z)
= \mathcal{L}_A(G,D_Y) + \mathcal{L}_A(F,D_X) \\
&+ \lambda_1\mathcal{L}_A(E,D_Z) + \lambda_2\mathcal{L}_{CX}(G,F) + \lambda_3\mathcal{L}_{CY}(G,F) \\
&+ \lambda_4\mathcal{L}_{CZ}(E,F),
\label{eq_totalloss}
\end{split}
\end{align}
where $\lambda_*$ controls the relative importance of different losses.w
We aim to solve:
$$
\footnotesize
G^*, F^*,E^*=\arg \min_{G,F,E} \max_{D_X,D_Y,D_Z} \mathcal{L}(G,F,E,D_X,D_Y,D_Z).
$$

This framework naturally fits the situation of asymmetric domains of different complexity, for we construct another path for information flow.
Moreover, since $\hat{X}$ is generated from both $Y$ and $Z$, the mapping $F$ is no longer injective and conditioned only on the input sample $y$, which enable us to obtain diverse $\hat{x}$ by tuning $z$ or utilizing $\hat{z}=E(x)$ that is encoded by a specific $x$ to control the generated mode.
Finally, since we continuously inject sampled $z$ during training, we are more likely to decouple the strong correlation between $G$ and $F$ to keep the final solution from being severely ill-conditioned and sensitive which also benefits the generation quality.

\subsubsection{Extensive Losses}
\label{subsec_ext}
Using the full objective in Sec.\ref{full_objective} is enough to train a reasonable model on relatively large dataset such as Cityscapes which owns thousands of training images for both domains.
However, in situations where the datasets are relatively small, or the information volume of two domains differs too much (such as edge$\leftrightarrow$photo), the objective in Eq.\ref{eq_totalloss} turns out to be not stable enough, thus causing a convergence or mode collapse problem.
The cycle consistency loss of $z$, $\mathcal{L}_{CZ}(E, F)$, is hardly able to converge.
Thus, to stabilize training and avoid mode collapse on small datasets such as day-night and vangogh2photo, we introduce some training techniques and extensive losses.

First, in the forward loop, we not only use $\hat{z}$ along with $\hat{y}$ to recover $x^*$, but also use the sampled auxiliary variable $z\sim p(z)$ to generate ${\hat x}_1$.
Since the sampled $z$ has nothing to do with $x$, it will be improper to require ${\hat x}_1 \approx x$.
Thus supervision is provided by $D_X$, as:
\begin{align}
\begin{split}
\mathcal{L}_{A}^{\mathrm{ext}}(G, F, D_X)&=\mathbb{E}_{x\sim p_{data}(x),z\sim p(z)}[\log D_X(x) \\
&+ \log (1-D_X(F(G(x),z)))]. \\
\end{split}
\end{align}
Similarly, in the backward loop, in addition to sampled $z$, we also use generated $\hat{z}$ to synthesize $\hat{x}_2$, which again need to be supervised by $D_X$:
\begin{align}
\begin{split}
\mathcal{L}_{A}^{\mathrm{ext}}(F,E, D_X)&=\mathbb{E}_{x\sim p_{data}(x),y\sim p_{data}(y)}[\log D_X(x) \\
& + \log (1-D_X(F(y,E(x))))]. \\
\end{split}
\end{align}
In this configuration, $\hat{x}_2$ should possess some residual information such as the style of $x$ while holding the content of $y$, which motivates us to introduce the perception loss for content and style in style transfer \cite{johnson2016perceptual} to enhance our control for output diversity.
A pre-trained network $\phi$ is used to evaluate the perception similarity of two images.
Let $\phi_j(x)$ be the activations of the $j$th layer of $\phi$ when processing the image $x$ whose shape is $C_j\times H_j\times W_j$, and the perception loss for content is :
\begin{align}
\begin{split}
\mathcal{L}_{content}^{\mathrm{ext}}(F,E)=&\frac{1}{C_j H_j W_j} \mathbb{E}_{x\sim p_{data}(x),y\sim p_{data}(y)} \\
& ||\phi_j(F(y, E(x)))-\phi_j(y)||^2_2,\\
\end{split}
\end{align}
and the perception loss for style is:
\begin{align}
\begin{split}
\mathcal{L}_{style}^{\mathrm{ext}}(F,E)
=&\mathbb{E}_{x\sim p_{data}(x),y\sim p_{data}(y)} \\
& ||P^\phi_j(F(y, E(x)))-P^\phi_j(x)||^2_F,\\
\end{split}
\end{align}
where $P^\phi_j(x)$ is a $C_j\times C_j$ Gram matrix \cite{johnson2016perceptual} whose elements are given by
\begin{align}
\begin{split}
P^\phi_j(x)_{c,c'}=
&\frac{1}{C_j H_j W_j} \sum_{h=1}^{H_j} \sum_{w=1}^{W_j} \phi_j(x)_{h,w,c} \phi_j(x)_{h,w,c'} .
\end{split}
\end{align}
Also, to encourage spatial smoothness of the generated images and reduce spike artifacts, total variation loss is introduced to regularize all the synthesis results:
\begin{align}
\begin{split}
\mathcal{L}_{TV}^{\mathrm{ext}}(G,F,E) =
&\mathbb{E}_{x\sim p_{data}(x),y\sim p_{data}(y),z\sim p(z)} \\
& [\varphi(G(x)) + \varphi(F(G(x),E(x))) \\
& + \varphi(F(G(x),z)) + \varphi(F(y,z))\\
& + \varphi(G(F(y,z))) + \varphi(F(y,E(x)))],
\end{split}
\end{align}
where$\varphi(x)= \sum_{h=1}^{H_j-1} \sum_{w=1}^{W_j-1} ((x_{w,h+1}-x_{w,h})^2-(x_{w+1,h}-x_{w,h})^2)$.

Therefore, the extensive loss version of full objective is as follows:
\begin{align}
\footnotesize
\begin{split}
&\mathcal{L}^{\mathrm{ext}}(G,F,E,D_X,D_Y,D_Z)
=\mathcal{L}_A(G,D_Y) + \mathcal{L}_A(F,D_X) \\
&+ \lambda_1\mathcal{L}_A(E,D_Z) + \lambda_2\mathcal{L}_{CX}(G,F) + \lambda_3\mathcal{L}_{CY}(G,F) \\
&+ \lambda_4\mathcal{L}_{CZ}(E,F) + \lambda_5\mathcal{L}_{A}^{\mathrm{ext}}(G, F, D_X) +  \lambda_6\mathcal{L}_{A}^{\mathrm{ext}}(F,E, D_X) \\
&+ \lambda_7 \mathcal{L}_{content}^{\mathrm{ext}}(F,E) + \lambda_8 \mathcal{L}_{style}^{\mathrm{ext}}(F,E) + \lambda_9 \mathcal{L}_{TV}^{\mathrm{ext}}(G,F,E),
\label{eq_totalloss_ext}
\end{split}
\end{align}
where $\lambda_*$ controls the relative importance of different losses.

These extension losses can be divided into 3 categories. 
$\mathcal{L}_{A}^{\mathrm{ext}}(G, F, D_X)$ and $\mathcal{L}_{A}^{\mathrm{ext}}(F, E, D_X)$ are adversarial losses.
$\mathcal{L}_{style}^{\mathrm{ext}}$ and $\mathcal{L}_{content}^{\mathrm{ext}}$ are perception losses.
$\mathcal{L}_{TV}$ loss is another kind.
Adding these losses to the full objective benefits model convergence and ensures $E$ to capture the lost visual representation dimensions of $X\rightarrow Y$ when training on small datasets, thus enhancing our control of the output diversity.
We will show the effect of these 3 kinds of losses respectively in Sec.\ref{subsec_abla}.

\vspace*{-15pt}
\section{Experiments}
\label{sec_experiment}

We conduct a series of experiments on Cityscapes \cite{cordts2016cityscapes}, Helen \cite{smith2013exemplar} datasets on label$\leftrightarrow$photo task to show our improvement in the aspect of generation quality, output diversity, and model sensitivity.

Cityscapes dataset is a semantic segmentation dataset consisting of 2975 training images and 500 validation images with pixel level annotations.
We conduct experiments of label$\leftrightarrow$photo task on Cityscapes dataset because domain label and domain photo are asymmetric, and more importantly, the results of both directions can be evaluated quantitatively, thus leading to an easy comparison with other methods.
Images in Cityscapes are actually paired.
We ignore the paired information during training and only use it in evaluation.
Helen is a dataset for face parsing with 2000 training images and 100 validation images.

According to our experiments, we want to show that: (1) our proposed AsymGAN is able to generate images of better quality under the circumstances where two domains are not symmetric; (2) in translations where the mapping should be one-to-many, we are able to generate diverse and reasonable output images by sampling different $z$, and control the output diversity by using $\hat{z}$ encoded by some other $x$; (3) what sensitivity problem is, and we alleviate this problem since we decouple the correlations of $G$ and $F$ by adding noise during training.
Additionally, we will discuss the influences of different losses and implementations.

\vspace*{-15pt}
\subsection{Implementation}
\label{subsec_implementation}
We adopt the same network architecture as \cite{zhu2017unpaired} for $G$,$D_X$ and $D_Y$.
The generator networks contain 2 stride-2 convolutions, 6 residual blocks \cite{he2016deep}, and 2 fractionally-strided convolutions with stride 1/2.
The network of $E$ contains 3 stride-2 convolutions and 3 residual blocks.
The generator $F$ takes $z$ and $x$ as input.
Based on the structure of $G$, we upsample $z$ to a proper size so that it can be concatenated to the middle feature map of network $F$, which is after the $3^{rd}$ residual block.
For the discriminator networks, we use $70\times 70$ PatchGAN \cite{zhu2017unpaired}.
The network of $D_Z$ is $20\times 20$ PatchGAN.
To stabilize training, we use LSGAN \cite{mao2017least} instead of the original negative log likelihood GAN for adversarial losses.
We set $\lambda_1=1, \lambda_2=\lambda_3=\lambda_4=10$ in Eq.\ref{eq_totalloss}.
Learning rate keeps 0.0002 and 0.0001 for $G$ and $D$ for 100 epochs and decays to zero over the next 100 epochs.
We represent this no extension loss version as "\emph{Ours-w/o-ex}" in all the experiments.

For the extensive loss version, besides the above settings, we further add an extra average pooling layer after the original $E$ to shrink $z$ to an 8-dimensional vector and add $z$ to the middle 3 residual blocks with Conditional Instance Normalization (CIN) \cite{dumoulin2017learned, perez2017film, Almahairi2018AugmentedCL}.
We set $\lambda_5 = \lambda_6 = 1, \lambda_7 =0.2, \lambda_8 = 0.1, \lambda_9 = 10 $ in Eq.\ref{eq_totalloss_ext}.
All these hyper-parameters are set empirically to make all the losses in the same order of magnitude and contribute almost the same to the gradient computing.
This extensive loss version is represented as ``\emph{Ours}" in all the experiments. 

We compare our method with DualGAN \cite{yi2017dualgan}, AugCGAN \cite{Almahairi2018AugmentedCL}, and Cycle GAN \cite{zhu2017unpaired}.
We train DualGAN\footnote{https://github.com/duxingren14/DualGAN}, 
AugCGAN\footnote{https://github.com/aalmah/augmented\_cyclegan}, 
and Cycle GAN\footnote{https://github.com/junyanz/pytorch-CycleGAN-and-pix2pix} models with their default settings and provided code on Cityscapes and Helen datasets.

\begin{table}
\centering
\normalsize
\begin{tabular}{l|ccc}
  \toprule
    & \textbf{Per-pixel} & \textbf{Per-class} & \textbf{Class}\\
  \textbf{Model} & \textbf{acc.} & \textbf{acc.} & \textbf{IOU}\\
  \midrule
  CoGAN \cite{liu2016coupled} & 45.0\% & 11.0\% & 8.0\% \\
  SimGAN \cite{shrivastava2016learning}  & 47.0\% & 11.0\% & 7.0\% \\
  DualGAN \cite{yi2017dualgan} & 49.3\% &11.2\% & 7.6\% \\
  AugCGAN \cite{Almahairi2018AugmentedCL} & 52.3\% & 17.8\% & 13.3\% \\
  Cycle GAN \cite{zhu2017unpaired} & 58.0\% & 22.0\% & 16.0\% \\
  CycleGAN baseline & 58.4\% & 21.2\% & 15.7\% \\
  \textbf{Ours-w/o-ex} & \textbf{74.9\%} & \textbf{27.6\%} & \textbf{21.6\%} \\
  Ours & 63.1\% & 22.1\% & 16.1\% \\
  \bottomrule
\end{tabular}
\caption{Segmentation scores on Cityscape photo$\rightarrow$label.}
\vspace*{-15pt}
\label{table_seg}
\end{table}

\begin{table}
\centering
\normalsize
\begin{tabular}{l|ccc}
  \toprule
    & \textbf{Per-pixel} & \textbf{Per-class} & \textbf{Class}\\
  \textbf{Model} & \textbf{acc.} & \textbf{acc.} & \textbf{IOU}\\
  \midrule
  \emph{FCN-8s} \cite{chen2016deeplab}  & & & \emph{65.3}\% \\
  \midrule
  \emph{FCN-8s-256} & \emph{80.0\%} & \emph{26.0\%} & \emph{21.2\%} \\
  CoGAN \cite{liu2016coupled} & 40.0\% & 10.0\% & 6.0\% \\
  SimGAN \cite{shrivastava2016learning} & 20.0\% & 10.0\% & 4.0\% \\
  DualGAN \cite{yi2017dualgan} & 45.9\% &11.2\% & 6.6\% \\
  AugCGAN \cite{Almahairi2018AugmentedCL} & 42.2\% & 15.6\% & 9.3\% \\
  Cycle GAN \cite{zhu2017unpaired}  & 52.0\% & 17.0\% & 11.0\% \\
  CycleGAN baseline & 53.9\% & 17.7\% & 12.6\% \\
  \textbf{Ours-w/o-ex} & \textbf{64.0\%} & \textbf{24.0\%} & \textbf{16.9\%} \\
  \midrule
  \emph{Our FCN-32s-256} & \emph{87.9\%} & \emph{70.9\%} & \emph{59.4\%} \\
  AugCGAN \cite{Almahairi2018AugmentedCL} & 46.8\% & 17.0\% & 11.4\% \\
  CycleGAN baseline & 60.9\% & 24.4\% & 17.9\%  \\
  Ours-w/o-ex & 74.3\% & 30.4\% & 22.9\% \\
  \textbf{Ours} &  \textbf{79.4\%} & \textbf{31.4\%} & \textbf{24.5\%} \\
  \bottomrule
\end{tabular}
\caption{FCN scores on Cityscape label$\rightarrow$photo.}
\vspace*{-5pt}
\label{table_fcn}
\end{table}

\begin{table}
  \centering
  \normalsize
  \begin{tabular}{l|ccc}
  \toprule
               & Correctness & Realness & Richness \\
  \midrule
  CycleGAN baseline & 2.91        & 2.85     & 2.87     \\
  Ours-w/o-ex & 3.62 & \textbf{3.51}& 3.47     \\
  \textbf{Ours}& \textbf{3.74}& 3.48 & \textbf{3.52}     \\
  \bottomrule
  \end{tabular}
  \caption{MOS of CycleGAN and AsmyGAN. 1 for \emph{Bad}, 2 for \emph{Poor}, 3 for \emph{Fair}, 4 for \emph{Good}, and 5 for \emph{Excellent}}
  \vspace*{-15pt}
  \label{tab_humaneval}
\end{table}

\vspace*{-15pt}
\subsection{Generation Quality}
The generation quality evaluation has always been a big problem in researches of GANs.
To evaluate both sides of the generation, we follow \cite{zhu2017unpaired} to conduct a photo$\leftrightarrow$label experiment on Cityscapes \cite{cordts2016cityscapes} and analyze the result both quantitatively and qualitatively.
Furthermore, we survey the generation quality with human questionnaires.
Sampling configuration is used for $Y\rightarrow X$ inference.

\subsubsection{Qualitative Evaluation}
Fig.\ref{fig_quality} shows our comparison with DualGAN \cite{yi2017dualgan}, AugCGAN \cite{Almahairi2018AugmentedCL}, and Cycle GAN \cite{zhu2017unpaired}.
Though sharing the same framework with Cycle GAN, the generator network of DualGAN is too simple to produce reasonable results for label$\leftrightarrow$photo task.
It generates the same output in both forward and backward loops, thus fails to model the mappings between label and photo domains.

For label$\rightarrow$photo, the results of Cycle GAN look bad in details; in line 2 and 4, the roads, trees, and cars are a mass of gray or green.
Surprisingly, the results of AugCGAN are also unsatisfactory. 
We find it in the next subsection that AugCGAN performs very well at multi-modality mappings, but its generation quality is poor.
The generated photos seem to be flat and just filled with colors, and the illumination and details are not appropriate.
Our results express more details like the lane line on the road, the windows, lamps, and shadows on the cars, and the texture of trees.
Second, Cycle GAN and AugCGAN models easily mix the categories up.
Especially, Cycle GAN turns over \emph{vegetation} and \emph{building} very often, like in line 1, 3, and 4.
In Line 2, it also confuses \emph{rail track} and \emph{sidewalk}.
AugCGAN just does not distinguish \emph{vegetation} and \emph{building} categories and generates categories randomly.
Our model distinguishes these categories clearly, which means that our model learns to understand the semantic information better.

For photo$\rightarrow$label, we also obtain visibly better results. 
AugCGAN and Cycle GAN models seem to learn less semantic mapping relationships during the unpaired training process.
Again, Cycle GAN confuses \emph{vegetation} and \emph{building} in line 1, and \emph{rail track} and \emph{sidewalk} in line 2.
Meanwhile, it is more likely to produce inconsistent and discontinuous labels for one large object, such as the big car in line 4.
AugCGAN just generate \emph{vegetation} and \emph{building} randomly.

\subsubsection{Quantitative Evaluation}
We first compare both sides of generation of \emph{Ours-w/o-ex} with CoGAN \cite{liu2016coupled}, SimGAN \cite{shrivastava2016learning}, DualGAN \cite{yi2017dualgan}, AugCGAN \cite{Almahairi2018AugmentedCL}, and Cycle GAN \cite{zhu2017unpaired}, and then compare the results of \emph{Ours-w/o-ex} with \emph{Ours} to see the influence of extensive losses on generation quality.

\begin{figure}
  \begin{center}
    \includegraphics[width=\linewidth]{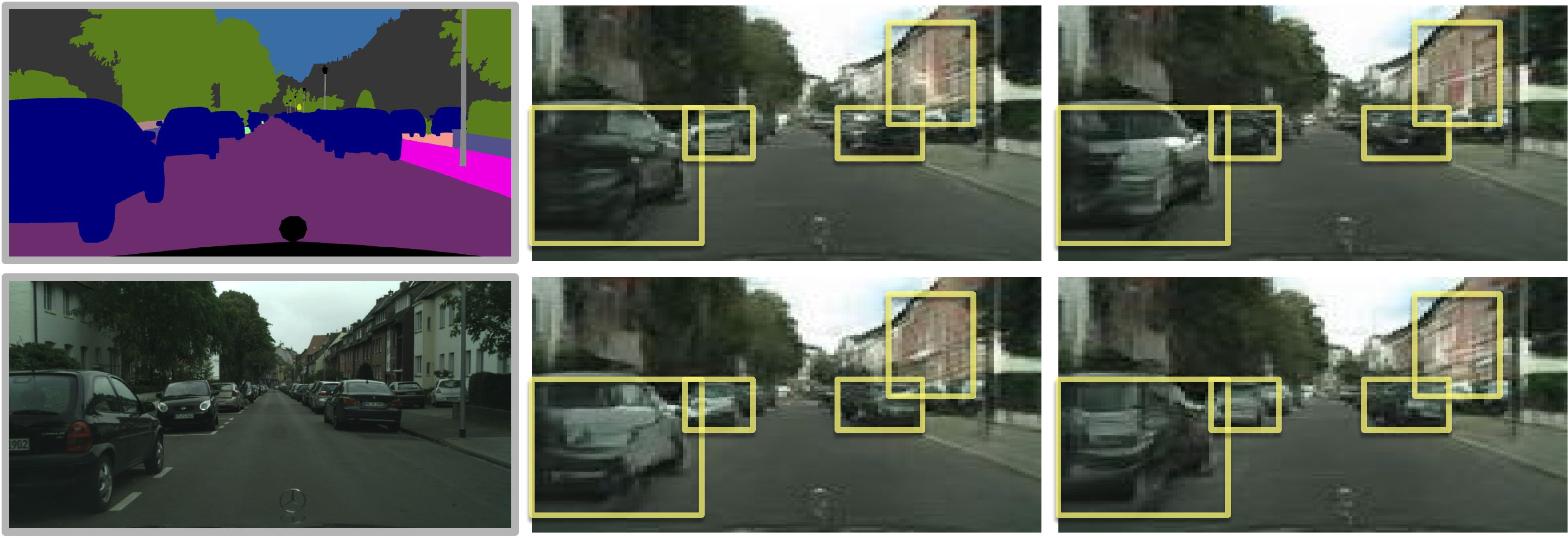}
    \caption{The generation results of $(Y,Z)\rightarrow X$ with different sampled $z$ on Cityscapes dataset. The first column is the input $y$ and its corresponding ground truth $x$. The other images are generated with different $z$. The diversity of this dataset is mainly represented by the color of cars or the texture of trees, which is not very obvious. We highlight the areas that differs from each other evidently.}
    \vspace*{-20pt}
    \label{fig_diversity_city}
  \end{center}
\end{figure}

\begin{figure*}
  \begin{center}
    \includegraphics[width=\linewidth]{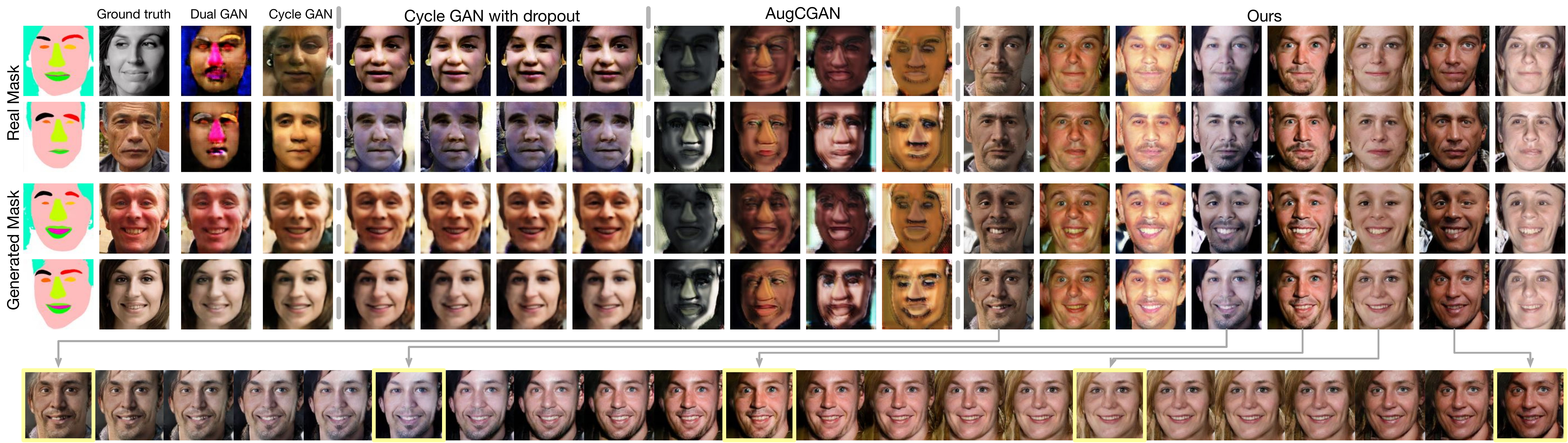}
    \caption{A diversity comparison with DualGAN \cite{yi2017dualgan}, Cycle GAN \cite{zhu2017unpaired}, Cycle GAN with dropout \cite{zhu2017unpaired}, and AugCGAN \cite{Almahairi2018AugmentedCL}. The generation results of $(Y,Z)\rightarrow X$ with different sampled $z$ on Helen dataset. Each column shares the same $z$. The last row represents the results of interpolation on $z$.}
    \vspace*{-20pt}
    \label{fig_diversity}
  \end{center}
\end{figure*}

\begin{figure*}
  \centering
  \begin{minipage}{0.48\linewidth}
  \centering
  \includegraphics[width=\linewidth]{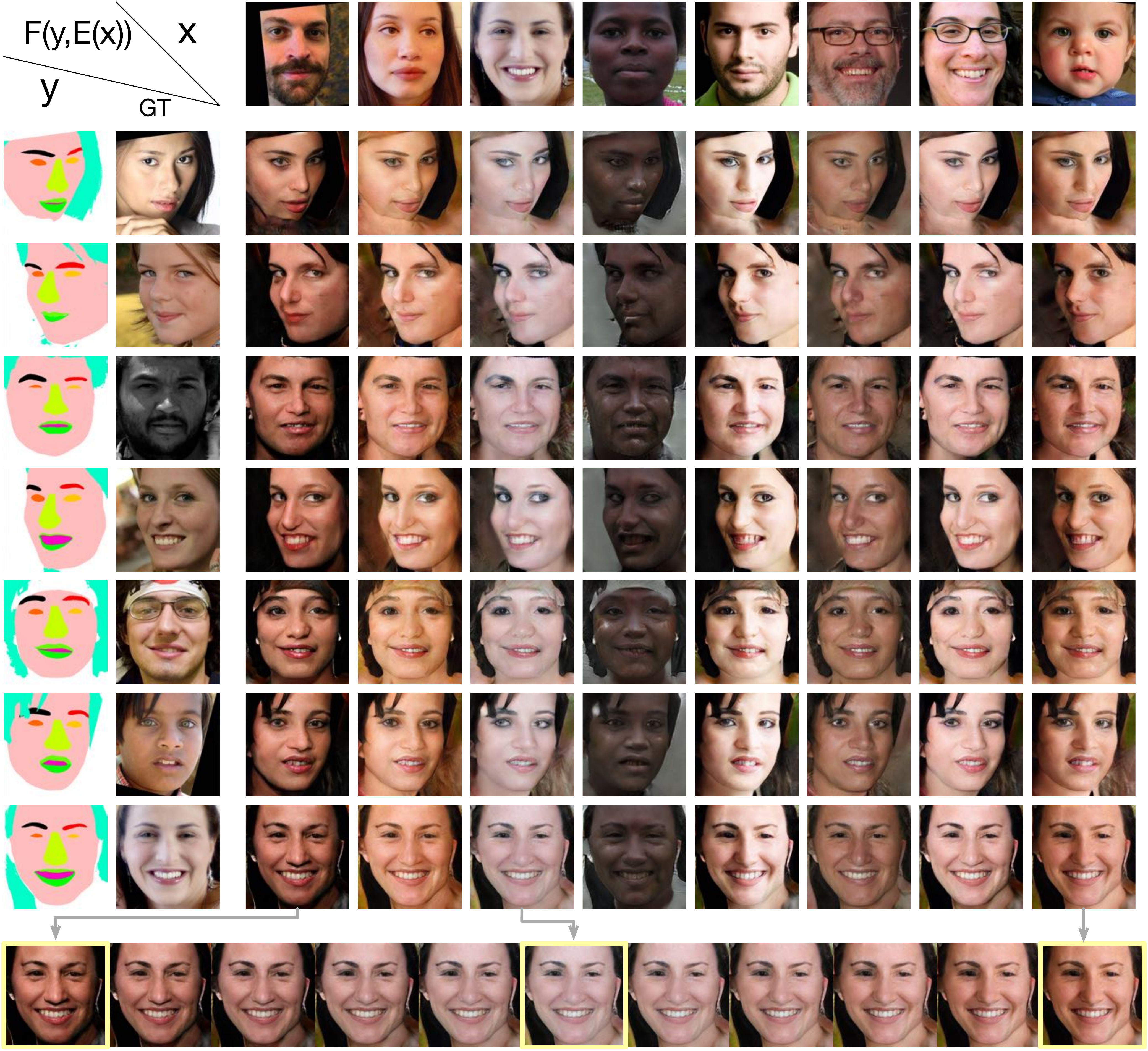}
  \caption{The generation results of $(Y,E(X'))\rightarrow X$ with different real $x$ on Helen dataset. The last row represents the results of interpolation on $\hat{z}$.}
  \label{fig_style_face}
  \end{minipage}
  \hfill
  \begin{minipage}{0.48\linewidth}
    \centering
    \includegraphics[width=\linewidth]{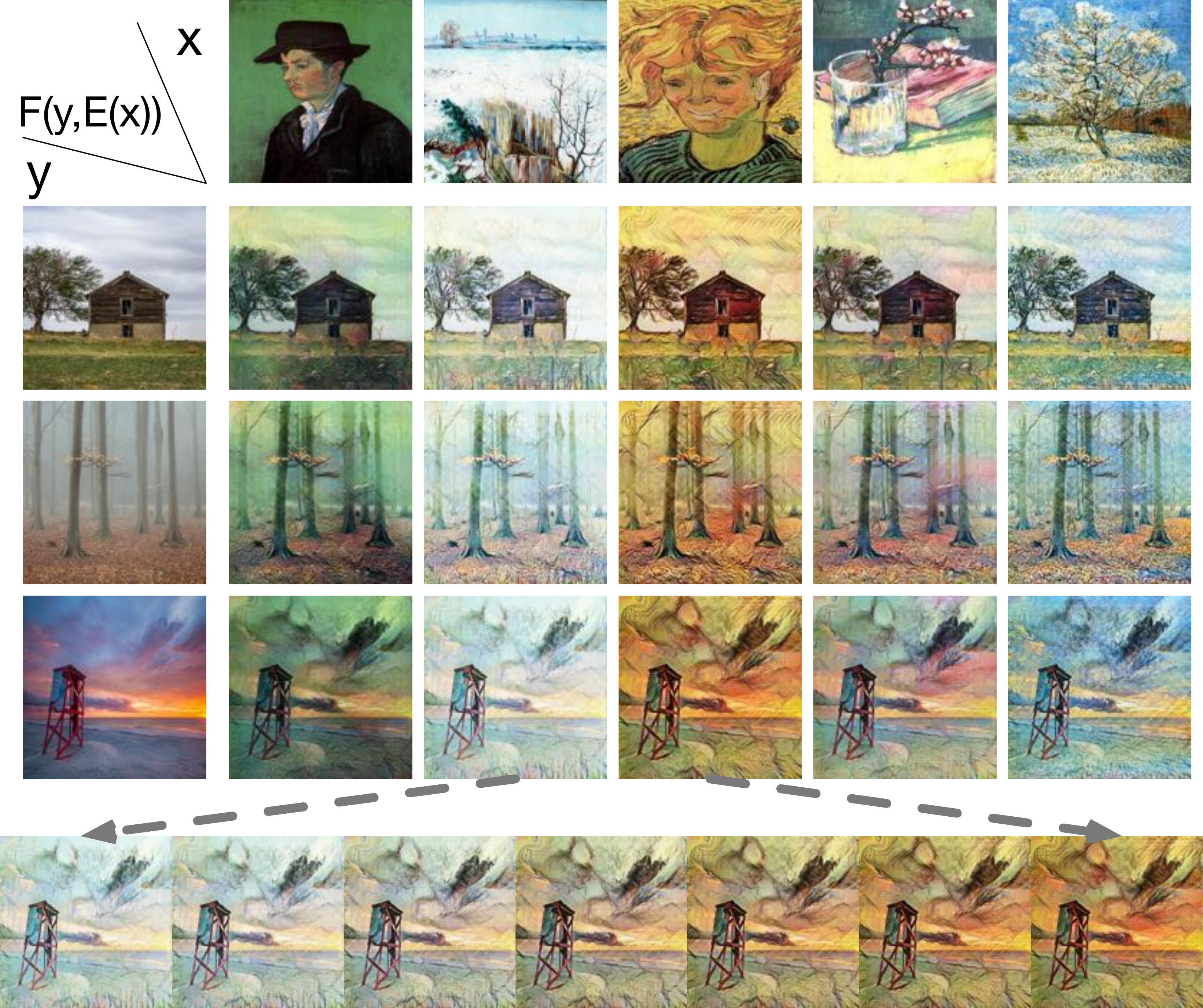}
    \caption{The generation results of $(Y,E(X'))\rightarrow X$ with different real $x$ on vangogh2photo dataset. The last row represents the results of interpolation on $\hat{z}$.}
    \label{fig_style}
  \end{minipage}
  \vspace*{-20pt}
\end{figure*}

For photo$\rightarrow$label, the ground truth label and segmentation metrics (per-pixel accuracy, per-class accuracy, and class IOU) can be directly used to evaluate the generation quality.
As shown in Table.\ref{table_seg},
we achieve the best performance over the other contrastive terms.
CycleGAN baseline denotes our trained Cycle GAN model, which reproduces the results given in \cite{zhu2017unpaired}.
Notably, \emph{Ours-w/o-ex} gains 16.5\%, 6.4\%, 5.9\% IOU improvement over CycleGAN baseline.

For label$\rightarrow$photo, FCN score \cite{isola2016image} is used for evaluation.
The main idea of evaluating the image quality is to segment the generated photos with a well-trained segmentation FCN model and calculate per-pixel accuracy, per-class accuracy, and class IOU, which are called FCN scores.
It is supposed that those who own better realness should be segmented better because the FCN model is trained for segmenting real photos, providing the upper bound of FCN scores.
Thus, we need to introduce a pre-trained FCN model to segment the generated fake images and compare the segmentation results with ground truth labels, as adopted in \cite{isola2016image}.
As shown in Table.\ref{table_fcn}, the FCN model (FCN-8s-256) that \cite{zhu2017unpaired} uses is trained on $256\times 256$ images whose class IOU on real validation images 21.2\% is far from 65.3\% reported in \cite{chen2016deeplab}.
Though FCN-8s-256 performs poor, for a fair comparison, we evaluate CycleGAN baseline and our results with FCN-8s-256 first.
Again, our trained CycleGAN baseline reproduces the results reported in \cite{zhu2017unpaired}, and Ours-w/o-ex outperforms CycleGAN baseline over 10.1\%, 6.3\%, and 4.3\%.
In addition, to make the evaluation more convincing, we train our own FCN model (\emph{Our FCN-32s-256}) with $256\times 256$ images, which achieves 59.4\% IOU on real images approaching to the reported 65.3\% \cite{chen2016deeplab} despite that our input resolution is lower than the original settings.
Thus, we evaluate CycleGAN baseline and our results with our FCN-32s-256.
Our model \emph{Ours-w/o-ex} still owns remarkably 13.4\%,6.0\%,5.0\% better performances on all the evaluation metrics than our reproduced CycleGAN baseline.
On the other hand, a significant gap still exists between our generated images and real images, thus indicating that translating images from information-poor domain to information-rich domain is still very challenging.

Comparing the results of \emph{Ours-w/o-ex} with \emph{Ours} for both directions, it should be noticed that adding extensive losses improves the quality of label$\rightarrow$photo while the performance of photo$\rightarrow$label drops a little.
It is because all the extensive losses are added to the label$\rightarrow$photo direction to enhance the generation quality of information-rich domain thus forcing the model to pay more attention to label$\rightarrow$photo that we are more interested in among the asymmetric domains' translations.
However, though the segmentation scores of \emph{Ours} for photo$\rightarrow$label are lower than that of \emph{Ours-w/o-ex}, they are still higher than the other state-of-the-art baselines.
Meanwhile, FCN scores of \emph{Ours} for label$\rightarrow$photo are even higher than that of \emph{Ours-w/o-ex}.

These experiments further verify that reserving the lost information to keep a balance between different complexity domains helps the model converge to a better solution, which improves the generation quality of both sides.

\subsubsection{Human Evaluation}
To further justify the generation quality of label$\rightarrow$photo, we conduct a human survey focusing on three attributes of generation quality: correctness of the generated categories concerning labels, realness of the whole photo, and richness of details.
We randomly sample 50 segmentation maps, and generate their corresponding photos with CycleGAN baseline and our method respectively to get 50 pairs of fake photos.
Then, we ask 45 people to rate these images.
For each participant, 20 fake photo pairs are presented in shuffled order.
These three attributes, correctness, realness, and richness, are scored with Absolute Category Rating \cite{itu1999subxjective} method on a scale ranging from 1 to 5 indicating Bad to Excellent by the participants.
Finally, we compute the mean opinion scores (MOS) for each attribute and show the results in Table.\ref{tab_humaneval}.
Generally speaking, in complex scenes such as Cityscape, the generation quality of CycleGAN baseline is between \emph{Poor} and \emph{Fair}, and our results are between \emph{Fair} and \emph{Good}.
Our method outperforms Cycle GAN in correctness, realness, and richness.
However, there is still great potential for future researches.

\vspace*{-15pt}
\subsection{Output Diversity}
The complexity asymmetry between domains always results in mapping ambiguous.
Thus, we should be able to generate several reasonable output images and provide a way to control the output diversity.
As expressed in Section \ref{sec_method}, we illustrate sampling configuration to generate diverse output images and encoding configurations to control the output diversity as shown in Fig.\ref{fig_framework}(4)(5).

\subsubsection{Sampled $z$}
First, we sample different \emph{aux}es $z\sim p(z)$ as in Fig.\ref{fig_framework}(4).
The results on Cityscapes datasets are shown in Fig.\ref{fig_diversity_city}.
Although the styles of the scene images in this dataset are almost the same in such a low resolution, we can still see that, as is highlighted, the color and texture of cars, buildings, and trees change with $z$.
To show the diverse output more intuitively, we train our model on another dataset with more variations, Helen \cite{smith2013exemplar}, a face parsing dataset consisting of 2k training images.
The human faces show more diversity than city streets.
The model on Helen dataset is trained with extensive loss version of the full objective in Eq.\ref{eq_totalloss_ext}.

We compare our results with DualGAN \cite{yi2017dualgan}, Cycle GAN \cite{zhu2017unpaired}, Cycle GAN with dropout, and AugCGAN \cite{Almahairi2018AugmentedCL}.
As discussed in Sec.\ref{subsec_aux}, Cycle GAN takes out the noise term $z$ due to its mode collapse problem.
To show this phenomenon more clearly, we run a Cycle GAN model with dropout, which performs dropout after each convolution layer both at training and inference phases.
The dropout version can be regarded as a simple implementation of adding $z$ as described in \cite{isola2016image, yi2017dualgan}.

The results are illustrated in Fig.\ref{fig_diversity}.
Again, DualGAN fails to model mappings between masks and photos.
Cycle GAN produces reasonable results, but it can only generate one output image.
Cycle GAN with dropout suffers from the mode collapse problem as claimed in \cite{zhu2017toward, Almahairi2018AugmentedCL}.
Since the generators are pushed to recover the original image no matter how the dropout performs, it is natural that the generators finally learn to ignore the noise term and produce almost the same outputs.
For AugCGAN and ours, we use the same sampled $z$ for each column on different input masks.

Though AugCGAN produces outputs with very high diversity, the generation quality is poor.
We think that the poor quality may be due to the fact that AugCGAN is designed to model many-to-many mappings between 2 domains where the information volumes are almost the same, like men$\leftrightarrow$women shows in \cite{Almahairi2018AugmentedCL} thus forcing $z_a$ and $z_b$ to model the differences between two domains explicitly.
For our focused asymmetric domains, \cite{Almahairi2018AugmentedCL} presents experiments on an edges$\leftrightarrow$shoes dataset which contains 50K images.
Therefore, on dataset such as Cityscapes and Helen that only contains 2-3K images, the results could be doubtful, and the bi-directional $z$es may introduce harmful disturbances for training.
Meanwhile for tasks like edges$\leftrightarrow$shoes, once the shoes are filled with some color, the output quality will seem good.
But for faces and city streets generation, we require more than filling the colors.

Comparing with AugCGAN, our method is able to generate diverse and high-quality outputs by tuning $z$.
The same $z$ represents similarly on different masks, and on both real and generated masks, which verifies that the model has learned to map features that are independent of segmentation mask to the latent space, and \emph{aux}es sampled from this latent space can present different features in generation.
Moreover, we conduct an interpolation experiment as is shown in the last row of Fig.\ref{fig_diversity}.
For any 2 sampled $z_s$ (\emph{e.g.} $z$ for the $1^{st}$ pic with yellow border) and $z_e$ (\emph{e.g.} $z$ for the $2^{nd}$ pic with yellow border), calculate the intermediate $z_i$s with linear interpolation, and generate the corresponding images $\{\hat{x_i} | i\in \mathbb{N}, 0<i<n, \hat{x_i}=F(y,z_i) \}$.
The interpolation results indicate that our learned $G$ and $F$ can build a continuous mapping between $Z$ and the photograph manifold.

\subsubsection{Encoded $z$}
Second, we encode different $\hat{z}$ from different $x$ to control the output diversity as illustrated in Fig.\ref{fig_framework}(5).
When we generate a face photo $\hat{x}=F(y, E(x))$, $\hat{x}$ is expected to contain the main content and structure of label $y$ and owns other features of face $x$.

\begin{figure*}
  \begin{center}
  \vspace*{-5pt}
    \includegraphics[width=\linewidth]{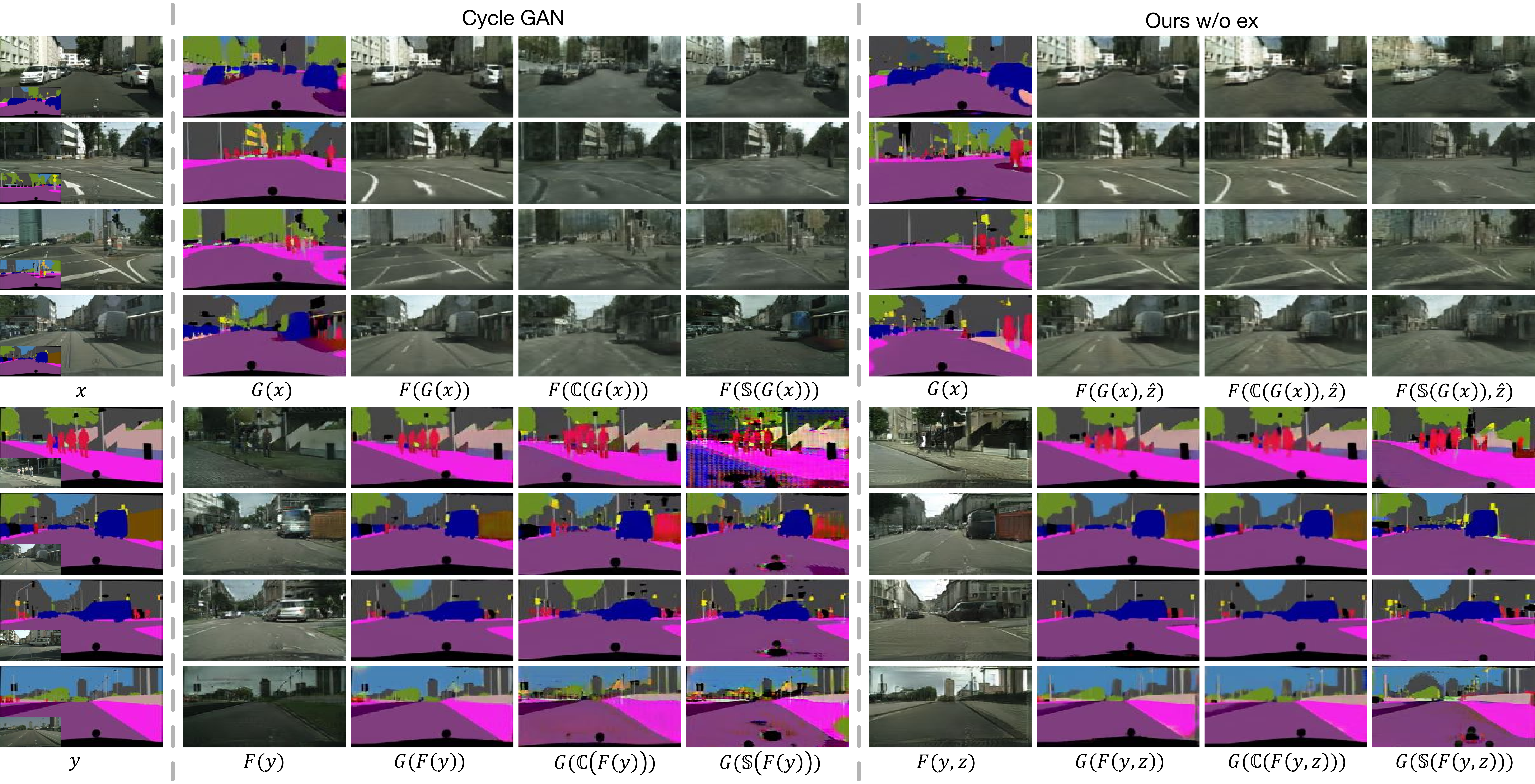}
    \caption{The influence of random crop $\mathbb{C}$ and scale variation $\mathbb{S}$ on CycleGAN baseline and our model. An illustration of sensibility problem. Our approach performs much better under the same circumstances.}
    \vspace*{-20pt}
    \label{fig_sensitivity_city}
  \end{center}
  \end{figure*}

\begin{figure}
\begin{center}
	\includegraphics[width=\linewidth]{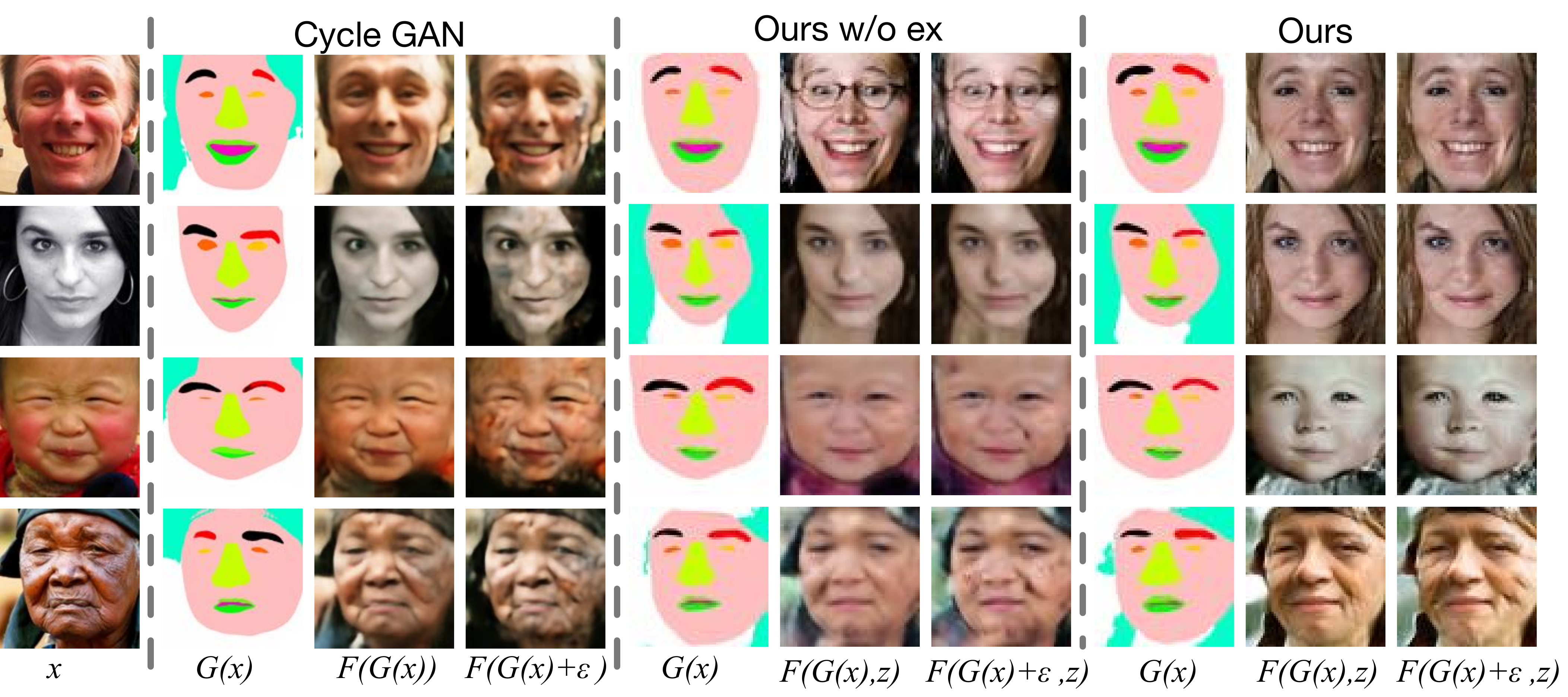}
	\caption{The influence of adding small disturbances $\epsilon$. An illustration of sensibility problem. Our approach performs much better under the same circumstances.}
	\vspace*{-20pt}
	\label{fig_sensitivity}
\end{center}
\end{figure}

As illustrated in Fig.\ref{fig_style_face}, we encode some testing images $x_i$ (the first row) to $\hat{z_i}$.
Then, generate faces with masks $y_j$ (the first column) and different $\hat{z_i}$.
As we can see, the model learns to control the color of faces, the color of mouses, and even the eyebrow shading from $x$.
The last row illustrates the results of interpolation.

To make this encoding configuration be more intuitively understood, we run our model on style transfer dataset vangogh2photo \cite{zhu2017unpaired}.
Results are shown in Fig.\ref{fig_style}
Though the information volumes of the style images' domain and real images' domain are relatively the same, with this experiments, we show that our model can not only be used in asymmetric domains but also be used to model multi-modality mappings.
We transfer a photo to Van Gogh's style owning a preference for a given painting in the aspect of color, hue, and texture.
The diversity introduced by color and hue is very intuitive.
For texture, the brushwork of the $1^{st}$ style image (green) is relatively smooth, the $3^{rd}$ (yellow) is thready, and the $5^{th}$ (blue) is punctate.
Take the forest content image as an example, when transferring it with the $1^{st}$ style image, the context of the generated image is quite smooth, and the $3^{rd}$ and $5^{th}$ generated paintings utilize more thready and punctate brushwork respectively corresponding to their style paintings.

These results show that our model successfully learns to control the output diversity.
Also, the interpolation results on $\hat{z}$ further verify that our encoder $E$ can map $X$ to $Z$ and that our modeled manifold is continuous.

\subsection{Model Sensitivity}

The model sensitivity problem is covert yet intriguing.
Analyzing the whole process of Cycle GAN, as shown in the first 3 columns in Fig.\ref{fig_sensitivity_city}, we observe that given a real $x$, no matter how $G$ work, $F(G(x))$ will always reconstruct $x$ almost perfectly.
In other words, generator $F$ works perfectly with generated fake masks $\hat{y}$, but with real masks $y$, $F(y)$ seems much worse and far from satisfied, no matter how similar $y$ and $\hat{y}$ are.
That is, $F$ performs differently with real masks and generated masks. 
(Also see Fig.\ref{fig_diversity} for different performances for real and fake masks of Cycle GAN and DualGAN.)
This phenomenon is inscrutable.
How can $F$ produce exactly the same real image when the output should have so many possibilities?
Why does $F$ perform so differently with generated masks and real masks?
How can $F$ recover the real image after losing so much information in masks?
Since this phenomenon shows even on validation set, it is not the network that remembers the input and output images, but $G$ and $F$ become reciprocal and coupled in some way.
We infer that maybe it is too hard to recover $x$ with $\hat{y}$, $G$  learn to encode the information of $x$ in $\hat{y}$ in some invisible ways and $F$ learns to exactly decode the information to recover $x$ to meet the cycle consistency loss.
The backward loop shows similar results.
However, because of the weird encoding schema, this reciprocal solution is very fragile.
We observe that the learned Cycle GAN model is sensitive to small location variation, small-scale variation, and small disturbances.
Furthermore, it is also because the generators pay too much attention to encoding and decoding information secretly, the generation quality is affected.

\subsubsection{The sensitivity problem}
The input $x$ is of size $128 \times 128$.
$F(G(x))$ reproduces $x$ perfectly.
First, we randomly crop $125 \times 125$ regions from $G(x)$ and generate $F(\mathbb{C}(G(x)))$, where $\mathbb{C}(i)$ denotes a random crop operation on image $i$, leading to notably worse results (the $4^{th}$ column of Fig.\ref{fig_sensitivity_city}).
Second, resizing $G(x)$ to $130 \times 130$ (denoting as $\mathbb{S}(G(x))$) also destroys the reciprocal condition of $G$ and $F$ and results in worse images (the $5^{th}$ column of Fig.\ref{fig_sensitivity_city}).
Third, we add some very small disturbances $\epsilon$ invisible to human to $G(x)$, the results $F(G(x)+\epsilon)$ again becomes seriously terrible.
Since Cityscapes images are relatively dim in illumination, the mass spots and patches cannot be observed obviously, we show the influence of small disturbances on Helen datasets as in the $4^{th}$ column of Fig.\ref{fig_sensitivity}.

As a result, we speculate that $G$ and $F$ encode the ``secret" information in small value of each pixel and the relative positions.
These values are too small to be observed.
Therefore, when we resize the images or adding disturbances to $G(x)$ or $F(y)$, we change its ciphertext, thus leading to bad decipher.
Similarly, maybe the starting point of decoding also matters a lot, so random cropping operation disorganizes decoding.

Such solutions of $G$ and $F$ are ill-conditioned and indicate that the network has converged to some trivial point that does harm to the generation quality of the models.

\subsubsection{Comparison}
In our framework, we continuously inject a noise term $z$ to the whole system, which breaking the coherence of $G$ and $F$, thus alleviating the sensitivity problem.
As illustrated in Fig.\ref{fig_sensitivity_city} and Fig.\ref{fig_sensitivity}, our model suffers less from both small location changes, scale changes, and disturbances.
In forward loop, random cropping leads to serious blurring and lacking local texture, which is notably worse than the original $F(G(x))$.
In contrast, although cropping and resizing still blur the output a little, our model performs much more robust.
$F(\mathbb{C}(G(x)), \hat{z})$ and $F(\mathbb{S}(G(x)), \hat{z})$ are still able to recover $x$, which indicates that our methods alleviate the coherence problem of $G$ and $F$.
Similarly, in backward loop, no matter how fake $F(y)$ looks like, Cycle GAN is able to predict a precise label for each pixel with $G(F(y))$, while with $G(\mathbb{C}(F(y)))$ and $G(\mathbb{S}(F(y)))$, the label maps become latticed and inconsistent.
On the other hand, our model shows strong robustness.
In fig.\ref{fig_sensitivity}, when we add very small disturbances $\epsilon$ invisible to human to $G(x)$, the results $F(G(x)+\epsilon)$ become spotted and stripy, while our model performs stably for both with and without extension losses versions.
Since $z$ is randomly sampled, $F(G(x), z)$ will not be precisely the save as $x$, but $F(G(x), z) \approx F(G(x)+\epsilon, z)$.

These experiments verify that our model is less sensitive to small location variations, scale variations, and disturbances than Cycle GAN, since $G$ and $F$ can be decoupled by the introduced asymmetric $z$, thus further ensures the network focusing on the generation.
Meanwhile, as we can see in Fig.\ref{fig_diversity}, our method performs almost the same for real and fake masks, which means our generator $F$ really learns to generate photos from masks but not secret encodings.

\begin{figure*}
  \begin{center}                         
    \includegraphics[width=\linewidth]{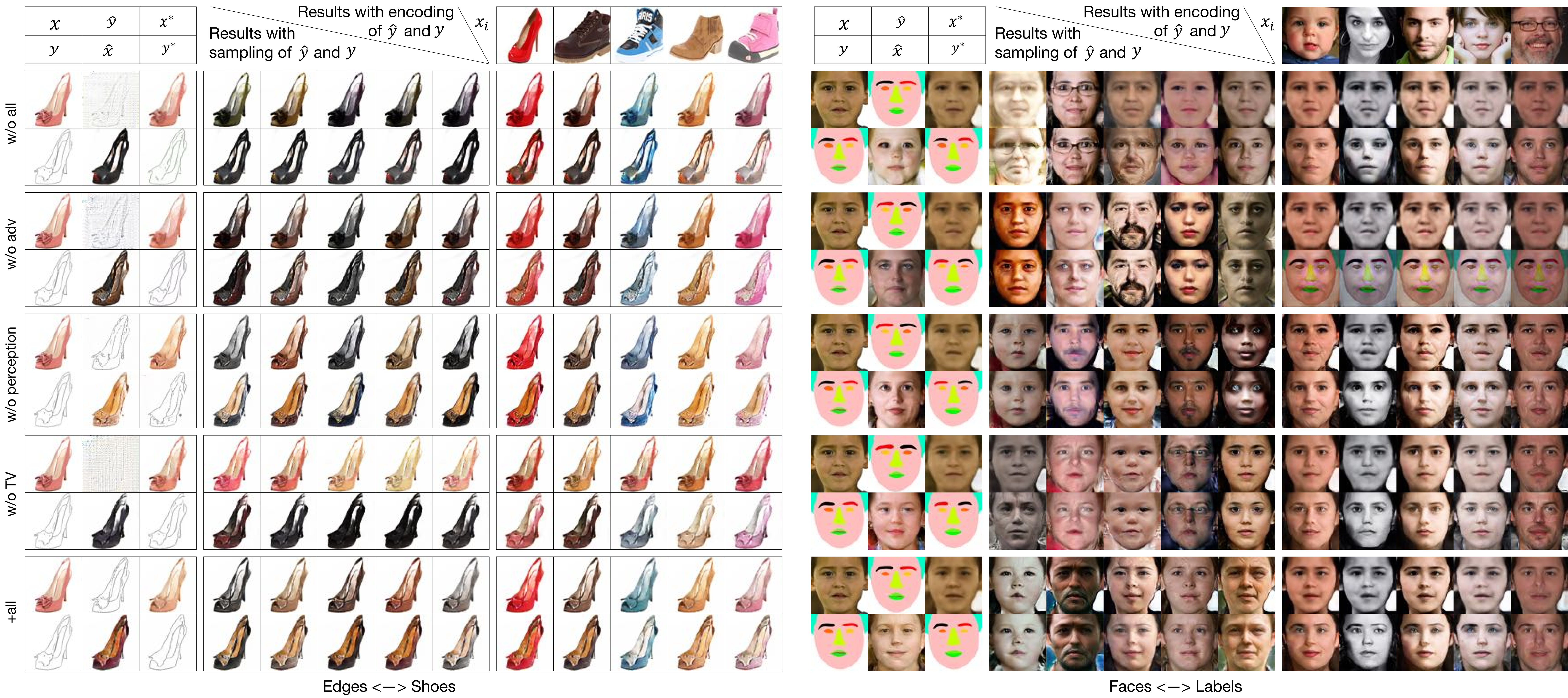}
    \caption{An ablation study about the effect of different extension losses. Left and right represent results of edges$\leftrightarrow$shoes and faces$\leftrightarrow$labels tasks.
	For each task, all rows are divided into 5 groups, corresponding to results of \emph{w/o all} (without all the extension losses), \emph{w/o adv} (without adversarial losses), \emph{w/o perception} (without perception losses), \emph{w/o TV} (without TV loss), and \emph{+all} (with all the extension losses).
	For each group, all the columns can be divided into 3 parts.
	The 1st row in part 1 shows results of $x\rightarrow \hat{y} \rightarrow x^*$; part 2 shows results of different sampled $z_i$ $\hat{x} = F(\hat{y}, z_i)$; part 3 shows results of encoded $\hat{z_i}$ corresponding to $x_i$ with $\hat{z_i} = E(x_i)$, $\hat{x} = F(\hat{y}, \hat{z_i})$.
	The 2nd row in part 1 shows results of $y \rightarrow \hat{x} \rightarrow y^*$; part 2 shows results of different sampled $z_i$ $\hat{x} = F(y, z_i)$; part 3 shows results of encoded $\hat{z_i}$ corresponding to $x_i$ with $\hat{z_i} = E(x_i)$, $\hat{x} = F(y, \hat{z_i})$.}
	\vspace*{-20pt}
    \label{fig_ablation}
  \end{center}
\end{figure*}

\begin{figure*}
  \begin{center}
    \includegraphics[width=\linewidth]{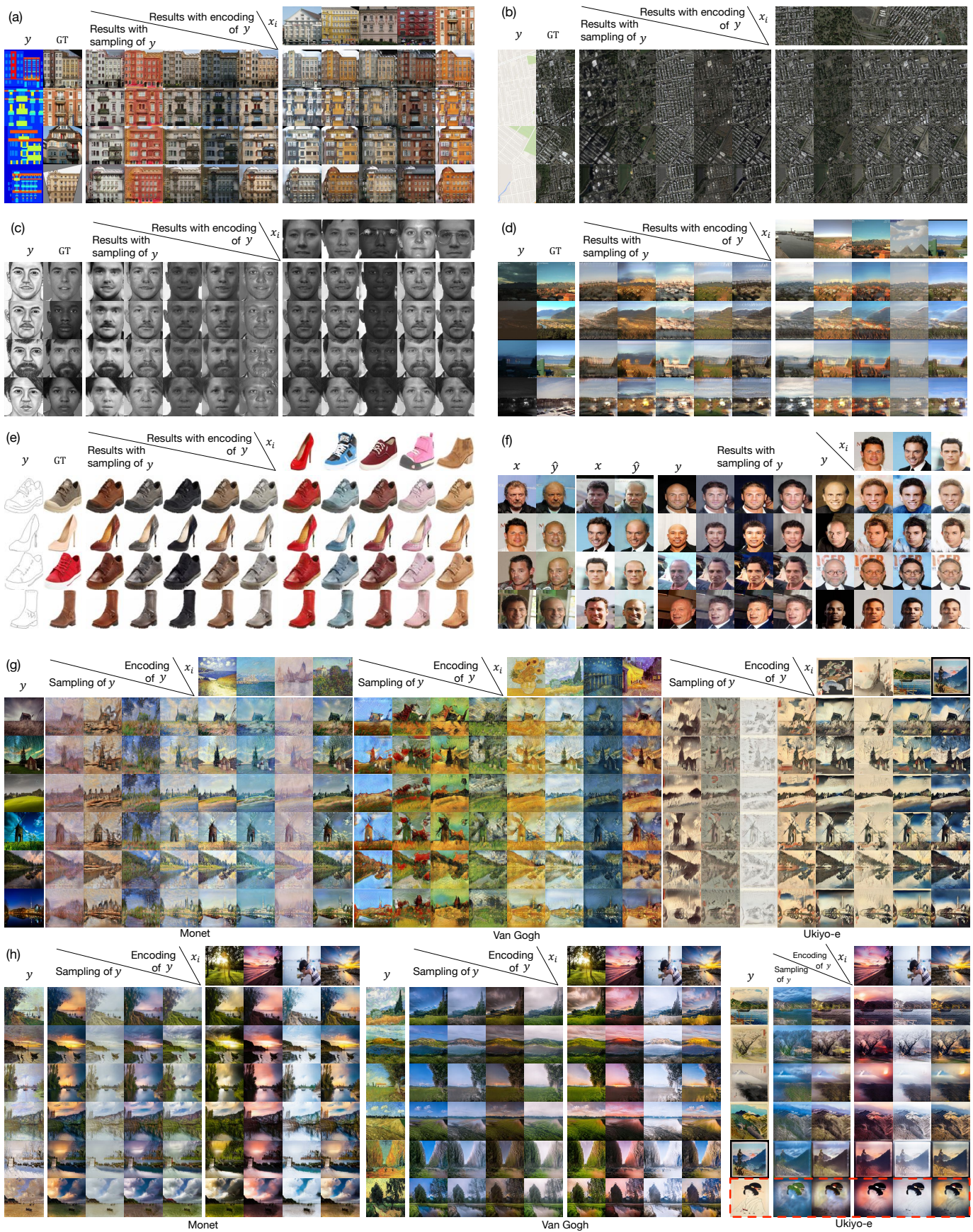}
    \caption{Applications. ``\emph{Results with sampling of $y$}" and ``\emph{Sampling of $y$}" represent results with sampling configuration of $y$ with $\hat{x} = F(y, z_i)$. $x_i$ represents encoding source images. ``\emph{Results with encoding of $y$}" and ``\emph{Encoding of $y$}" represents results with encoding configuration of $y$ with $\hat{x} = F(y, E(x_i))$. (a) Facades labels $\leftrightarrow$ photo; (b) Map $\leftrightarrow$ aerial photo; (c) Sketch $\leftrightarrow$ photo; (d) Night $\leftrightarrow$ day; (e) Edges $\leftrightarrow$ shoes; (f) Bald $\leftrightarrow$ hair; (g) Style transfer; (h) Painting $\rightarrow$ photo. The red box represents a failure case.}
    \label{fig_applications}
  \end{center}
\end{figure*}

\vspace*{-15pt}
\subsection{Ablation Studies}
\label{subsec_abla}

\paragraph{The effect of different extension losses}

As discussed in Sec.\ref{subsec_ext}, we add 3 kinds of extension losses to our main framework.
To illustrate the effect of these 3 kinds of extension losses respectively, we conduct a set of experiments on edges2shoes-3K and Helen datasets for edges$\leftrightarrow$shoes and label$\leftrightarrow$photo tasks with $64\times 64$ input images.
Especially, we sample 3K and 200 images from edges2shoes dataset \cite{yu2014fine} for both edges' domain and shoes' domain to build the training and validation sets of edges2shoes-3K dataset since the original edges2shoes dataset is too large and it takes a very long time to train on.
Results are shown in Fig.\ref{fig_ablation}.

We want our model to achieve the following goals:
(1) the model should generate reasonable results for both forward and backward loops;
(2) the generation quality should be good;
(3) the model should show diversity when inferencing with sampling configuration;
(4) the model should show diversity when inferencing with encoding configuration and the output diversity should correspond to the encoding image;
(5) the model should perform similarly on fake and real masks (or edges), so that the model is not learned to encoding and decoding secrets.

First, we train a model without all the extension losses with only the main framework.
We are able to get satisfactory results on some datasets such as cityscapes.
On Helen, though the generation quality is not very good, the results seem tolerable.
But on some tasks where the information volume of 2 domains are too different such as edges$\leftrightarrow$shoes, or the number of training images are too small such as day$\leftrightarrow$night and vangogh$\leftrightarrow$photo, convergency problems occur.
The main problems are that the model fails to pull the distributions of $\hat{y}$ and $y$ closely enough and fails to pull the distributions of sampled $z$ and encoded $z$ together.
As shown in Fig.\ref{fig_ablation}-\emph{w/o all}, $G$ fails to translate shoes to edges, and for edges$\rightarrow$shoes, it fails to produce diverse outputs with sampled $z$.

Second, the first kind of extension loss is adversarial losses added to $F(\hat{y}, z)$ and $F(y, \hat{z})$, while in our main framework, adversarial losses are only added to $F(\hat{y}, \hat{z})$ and $F(y, z)$).
The adversarial losses make $F(\hat{y}, z)$ and $F(y, \hat{z})$ less blur, especially among which $F(y, \hat{z})$ is our encoding configuration for inference which counts a lot.
Also, they provide more training data for $F$ and $D_X$, which improves the quality of $F$ and $G$ since they are highly correlated.
In Fig.\ref{fig_ablation}, comparing \emph{w/o Adv} with \emph{+all}, we can see that without adversarial losses, $G$ in edges$\leftrightarrow$shoes tasks fails again, and the encoding configuration in Helen also fails.
With adversarial losses, the faces are more clear and with better quality.

Then, the second kind of extension losses is perception losses.
These losses are applied only on the encoding configuration $F(y, \hat{z})$, thus having less influence on the other aspect.
However, as shown by the encoding configuration on Helen dataset, comparing results of \emph{w/o perception} with \emph{+all} and \emph{w/o TV}, it is found that perception losses help the model to keep more detailed features of the encoded source image, such as the color of eyes (the $1^{st}$ and $4^{th}$ columns of encoding), and the texture of faces (the last column of encoding).

The last extension loss is TV loss.
TV loss mainly works for ensuring the continuity of generated images and preventing grid-like fakes.
For edges$\leftrightarrow$shoes, without TV loss (\emph{w/o TV}), $G$ produces grid-like fakes, which fails the whole training, thus leading great difference between the generations of real and fake edges, even when applying encoding configuration.
On Helen dataset, comparing \emph{w/o TV} with the others that use TV loss, it is obvious that without TV loss, the generated faces tend to show some fake lines as is cracked, and with TV loss, the skins are smoother and more natural.

Above all, with all these extension losses, our model is able to achieve our goals and performs stable and robust on all the datasets we have tried including the following applications.

\paragraph{How to combine $z$ in $F$}

\begin{figure}
  \begin{center}
    \includegraphics[width=\linewidth]{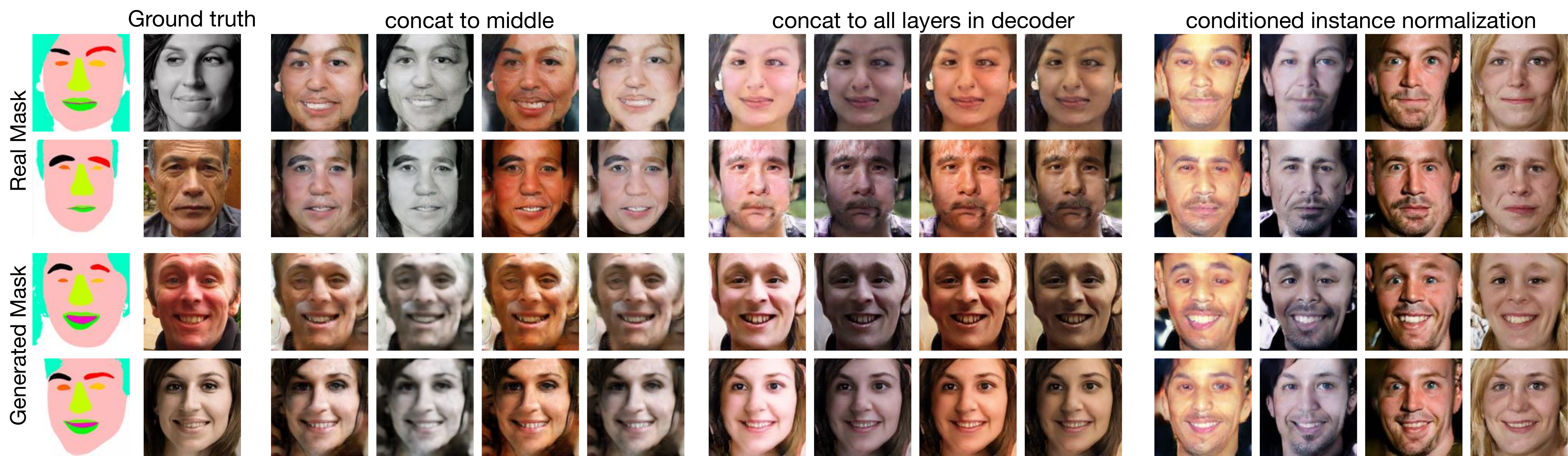}
    \caption{Ablation study about how to combine $z$ in $F$. The generation results of $(Y,Z)\rightarrow X$ with different sampled $z$ on Helen dataset. Each column shares the same $z$.}
    \vspace*{-20pt}
    \label{fig_ablation_addz}
  \end{center}
\end{figure}

How to design the structure of $F$ so that $z$ can introduce output diversity while preventing mode collapse is of great concern.
There are several options.
First, we upsample $z$ and concatenate it to the middle feature map of $F$.
Second, following \cite{zhu2017toward} we add an average pooling layer after the original $E$ to shrink $z$ to an 8-dimensional vector, and concatenate it to every feature map after the middle feature map of $F$, namely concatenate $z$ to all feature maps in decoder.
Third, motivated by \cite{Almahairi2018AugmentedCL}, we add $z$ to the middle 3 residual blocks with CIN to normalize these feature maps with scales and biases produced by $z$.
The results are shown in Fig.\ref{fig_ablation_addz}.
All the 3 approaches can introduce diversity to the output.
However, as we can see, concatenation related approaches mainly capture the diversity of all kinds of colors such as face and mouth, but other features are not changed very obviously.
CIN approach encodes more aspects of diversity such as beard and wrinkle.
As a result, we choose to use CIN in our framework.

\vspace*{-15pt}
\section{Applications}
\label{sec_app}
\vspace*{-5pt}
In this section, we illustrate 6 applications of asymmetric domains' image-to-image translations including architectural facades labels$\leftrightarrow$photos, map$\leftrightarrow$aerial photo, sketch$\leftrightarrow$photo, day$\leftrightarrow$night, edges$\leftrightarrow$shoes, and hair$\leftrightarrow$bald.
We also show style transfer applications and painting$\rightarrow$photo applications to verify the effectiveness of our method modeling not only asymmetric mappings but also multi-modality mappings between regular domains.
All results are shown in Fig.\ref{fig_applications}.
All these applications take $128\times 128$ images as input and produce output images at the same size.
We use the extension loss version for all these applications, and hyper-parameters and settings are described in Sec.\ref{subsec_implementation}.

\vspace*{-10pt}
\subsection{Architectural facades label$\leftrightarrow$photo: Fig.\ref{fig_applications}(a)}
We use the facade dataset \cite{tylevcek2013spatial} following \cite{zhu2017unpaired}.
This dataset contains 400 facade photos and labels.

\vspace*{-10pt}
\subsection{Map$\leftrightarrow$aerial photo: Fig.\ref{fig_applications}(b)}
Maps dataset \cite{isola2016image} contains 1096 training images scraped from Google Maps around New York.

\vspace*{-10pt}
\subsection{Sketch$\leftrightarrow$photo: Fig.\ref{fig_applications}(c)}
We run this task with PHOTO-SKETCH dataset \cite{wang2009face} following \cite{shen2017learning}.
This dataset contains 995 images for training and 199 for validation.

\vspace*{-10pt}
\subsection{Night$\leftrightarrow$day: Fig.\ref{fig_applications}(d)}
Day-night dataset \cite{laffont2014transient} contains only 100 images.
We follow \cite{shen2017learning} to use 90 images for training and 10 for validation.
Though the number of training images is very small, our model also produces satisfactory results.

\vspace*{-10pt}
\subsection{Edges$\leftrightarrow$shoes: Fig.\ref{fig_applications}(e)}
We samples only 3K images from Edges2shoes dataset \cite{yu2014fine} for fast training.

\vspace*{-10pt}
\subsection{Bald$\leftrightarrow$Hair: Fig.\ref{fig_applications}(f)}
We select 3K images from celebA dataset \cite{liu2015deep} with label ``Male=1" and ``Bald=0" for Hair domain, and ``Male=1" and ``Bald=1" for Bald domain respectively.
Select 200 images for both domains for validation.
As we can see, our model learns to turn a man with hair to bald.
Reversely, we can draw different hairstyles to bald men.
For the encoding configuration, we initially hope the model can hold the whole hairstyle of the encoding source images, but now we only capture the color of hair as a distinct factor while the hairstyle varies less.

The above applications are all image-to-image translation tasks between domains where the information volumes are quite different.
Our method shows high quality and diverse results for both sampling and encoding configurations on all the datasets without changing any hyper-parameters.

\vspace*{-10pt}
\subsection{Style transfer: Fig.\ref{fig_applications}(g)}
\label{subsec_styletransfer}
The earliest version of style transferring \cite{gatys2016image} model can only generate one image to one style.
Speedup versions \cite{johnson2016perceptual,li2016precomputed} have managed to learn a network to generate any content images to still one style image.
Until very recently, works like \cite{chen2017stylebank, zhang2018style} attempt to capture multiple styles in a single network and are able to transfer any content images to multiple styles.
These style transferring approaches mainly employ the perception loss for style and content to train model directly, without any adversarial process.
Additionally, Cycle GAN \cite{zhu2017unpaired} proposes to transfer a real photo to a style of a painter which is defined by a set of his paintings so that they utilize adversarial losses to make the transferred image indistinguishable from the domain of style paintings.
However, the style of a painter varies on different paintings in the aspect of color, hue, texture, and so on.
Different from theirs, our model is able to transfer any photos to any specific styles of a painter by combining the photo and the \emph{aux} encoded by another painting with encoding configuration.
Meanwhile, with sampling configuration, we can generate different styles of the artist without an explicit painting as a clue.
We run our model on Monet2photo, Vangogh2photo, and Ukiyoe2photo datasets \cite{zhu2017unpaired}.
These datasets share the same photo domains with landscape photographs downloaded from Flickr and WikiArt.
The artists' domains contain 1073, 400, 563 paintings for Monet, Van Gogh, and Ukiyo-e respectively.
We obtain a variety of transferring results with sampling and encoding configurations.

\vspace*{-10pt}
\subsection{Painting$\rightarrow$photo: Fig.\ref{fig_applications}(h)}
We use the same datasets as in Sec.\ref{subsec_styletransfer}.
Only makes the photo domain to be $X$ and painting domain to be $Y$, which is opposite from style transfer.
Since the photo domain mainly contains landscapes photos, which are similar to the paintings of Monet and Van Gogh, results on monet$\rightarrow$photo and vangogh$\rightarrow$photo show good quality.
However, the Ukiyo-e domain contains lots of portraits and humans.
Thus, when translating landscape paints, we can still get reasonable photos.
But when translating portrait painting, the model fails to turn the paintings to real photos, as shown in the last row of ukiyoe2photo.

The results of style transfer and painting$\rightarrow$photo show that our method can not only model mappings between asymmetric domains but also solve the multi-modality mappings between domains where the information volumes vary less.

% \vspace*{-10pt}
\section{Conclusion}
% \vspace*{-5pt}
In this paper, we propose an Asymmetric GAN approach for unpaired image-to-image translation focusing on the asymmetric situations.
We propose to introduce an auxiliary variable that follows a prior distribution to Cycle GAN framework so that we can (1) generate images with better quality since we try to reduce the loss of information by adding the path of \emph{aux};
(2) produce diverse target images by sampling and control the output diversity by encoding, because \emph{aux} allows us to model the target distribution instead of an injective mapping;
(3) alleviate the sensitivity problem in Cycle GAN due to the reason that we decouple $G$ and $F$ by continuously injecting \emph{aux} to $F$ during training, which further improves the generation quality.
Extensive experiments on Cityscapes, Helen and other datasets verify the effectiveness of our method both qualitatively and quantitatively.
Many applications on other asymmetric domains and multi-modality modeling tasks further show the robustness and generalization ability of our method.

\ifCLASSOPTIONcaptionsoff
  \newpage
\fi
% \vspace*{-20pt}
\bibliographystyle{IEEEtran}
\bibliography{AsymGAN-bib}

\end{document}